\documentclass[acmsmall, screen]{acmart}

\AtBeginDocument{%
  }

\usepackage{graphicx}
\usepackage{multirow}
\usepackage{amsthm}
\usepackage{mathrsfs}
\usepackage[title]{appendix}
\usepackage{xcolor}
\usepackage{textcomp}
\usepackage{manyfoot}
\usepackage{booktabs}

\usepackage{algorithm}
\usepackage{algorithmic}
\usepackage{listings}

\usepackage{subcaption}
\usepackage{xspace}
\usepackage{enumitem}
\usepackage{bm}
\usepackage{comment}

\theoremstyle{plain}

\theoremstyle{definition}

\theoremstyle{remark}

\newcommand{\ts}{time-series\xspace}
\newcommand{\framName}{\textbf{\texttt{CoMeTS-GAN}}\xspace}

\NewDocumentCommand{\grad}{e{_^}}{%
  \mathop{}\!% \mathop for good spacing before \nabla
  \nabla
  \IfValueT{#1}{_{\!#1}}% tuck in the subscript
  \IfValueT{#2}{^{#2}}% possible superscript
}

\setcopyright{acmlicensed}
\copyrightyear{2018}
\acmYear{2018}
\acmDOI{XXXXXXX.XXXXXXX}

\acmJournal{JDIQ}
\acmVolume{37}
\acmNumber{4}
\acmArticle{111}
\acmMonth{8}

\begin{document}

%%%% Short title for page headers
\title{High-Quality Synthetic Financial Time-Series using a GAN–Diffusion Framework}

\author{Giuseppe Masi}
\email{masi.g@di.uniroma1.it}
\affiliation{%
  \institution{Sapienza University of Rome}
  \country{Italy}
}

\author{Andrea Coletta}
\email{andrea.coletta@bancaditalia.it}
\authornote{The opinions expressed in this paper are personal and should not be attributed to Banca d'Italia.}
\affiliation{%
  \institution{Banca d'Italia}
  \country{Italy}
}

\author{Novella Bartolini}
\email{novella@di.uniroma1.it}
\affiliation{%
  \institution{Sapienza University of Rome}
  \country{Italy}
}

\renewcommand{\shortauthors}{Masi et al.}


% This special issue invites contributions that develop innovative quality metrics, report on relevant case studies, or propose novel SDG and DA methods that include quality guarantees by design.

\begin{abstract}
In recent years, financial institutions and firms have increasingly adopted synthetic data to address data scarcity and to generate counterfactual market scenarios.
However, reproducing all the statistical properties of financial time series, commonly known as stylized facts, remains an open challenge for many existing general-purpose architectures.
In this paper, we present a quality-aware generative framework that combines two classes of generative methods, demonstrating how their integration addresses existing limitations while enhancing the realism of synthetic data. Specifically, we first introduce \framName (\textbf{Co}rrelated \textbf{M}ultivariat\textbf{e} \textbf{T}ime \textbf{S}eries \textbf{GAN}), a Conditional Generative Adversarial Network (C-GAN) designed to jointly generate mid-price and volume \ts for correlated stocks.
We then show how our GAN architecture can be incorporated into state-of-the-art diffusion models to enhance the quality of generated correlation structures. Specifically, the GAN's Critic serves as a quality evaluation module that guides the diffusion process, enforcing learned correlation structures in the generated \ts.
Our framework offers a lightweight and responsive solution for realistic stock market simulation, explicitly modeling inter-asset correlation structures.
We experimentally validate our framework against leading generative architectures, showing that it more effectively captures the stylized facts of stock markets and models inter-asset correlations.
%
%The source code is publicly available on GitHub\footnote{GitHub repository: \url{https://github.com/giuseppemasi99/COMETS-GAN}.}.
\end{abstract}





\received{20 February 2007}
\received[revised]{12 March 2009}
\received[accepted]{5 June 2009}

\maketitle

\section{Introduction}
\label{sec:introduction}
%INTRO
The recent advancements in Artificial Intelligence (AI) have had a substantial impact on the financial industry~\cite{cao2022ai,ozbayoglu2020deep}, enabling novel algorithms for stock market prediction~\cite{prata2024lob}, risk assessment and management tools~\cite{mashrur2020machine}, portfolio hedging solutions~\cite{buehler2019deep}, or automated credit scoring systems~\cite{moscato2021benchmark}.
The effectiveness of AI in the financial domain is --- in part --- due to the fact that financial firms are early adopters of novel technologies, but it has also been largely driven by the huge amount of historical and high-frequency data.
In fact, financial markets are inherently data-driven and require precise, data-driven decision-making. 
Such an environment seems a natural fit for machine learning models since they 
%.Machine learning models 
could exploit large datasets to extract patterns, model complex dynamics, and improve predictive accuracy.

%\paragraph{Data availability}
However, the huge amount of data generated within financial systems does not always translate into available datasets for machine learning algorithms, especially for the academic community. 
This is primarily due to strict regulations --- which limit data sharing and usage --- as well as the high costs associated with high-frequency data sources. 
For example, while daily stock price data is generally accessible, more detailed data --- such as limit order book data --- is often proprietary and not publicly available. 
Additionally, rare and extreme market events are, by definition, underrepresented in historical data, posing novel challenges for training robust models or conducting thorough back-testing.

%SOLUTION
\paragraph{\textbf{Synthetic data}}
Synthetic data offers a promising solution to these challenges by providing data with the same characteristics and structure as real data, while avoiding the associated risks and constraints~\cite{potluru2023synthetic,jordon2022synthetic}.
Synthetic data tools enable the generation of \textit{theoretically} unlimited amounts of free data; they support the creation of counterfactual scenarios for testing and what-if analyses; and they allow data sharing without compromising privacy or exposing sensitive information.  
%EXISTING WORK
In recent years, a number of approaches have been designed to generate synthetic data for fundamental financial applications. 
Some notable examples include market simulations~\cite{colettaRealisticMarketSimulations2021a}, model calibration~\cite{cuchieroGenerativeAdversarialNetwork2020}, portfolio construction~\cite{papenbrockMatrixEvolutionsSynthetic2021, martiCORRGANSamplingRealistic2020}, design of hedging
strategies~\cite{cont2025data}, or counterfactual markets scenarios generation~\cite{coletta2023constrained}. 
In all these settings, developing data-generation techniques that accurately capture the complexities of global financial markets --- thus produce high-quality synthetic data --- is essential to the effectiveness of downstream tasks. 
Financial data have specific statistical properties, often referred to as stylized facts~\cite{cont2001empirical}, that characterize stock returns, volatility, orders, and all financial data. However, reproducing such properties is not a trivial task.
While recent work addresses the specific task of realistic financial \ts generation~\cite{tanaka2025cofindiff}, by conditioning the generative model to preserve the stylized facts, many statistical properties have remained overlooked, and quality-aware frameworks are still missing.
In particular, the correlation structure among multiple assets is often neglected, with only a few approaches addressing this issue~\cite{masi2023correlated}, thereby limiting the realism and applicability of the generated data in multivariate settings.

\paragraph{\textbf{The need for realistic correlated time-series}}
Correlation dynamics are among the most important statistical properties in the generated \ts. 
In the financial domain, correlation dynamics play an essential role in managing the risks within a portfolio by ensuring that the individual assets are not overly correlated with one another and impacted by similar market conditions~\cite{pafka2004estimated}.
There are several reasons why stocks may exhibit correlation.
Correlation may be sector-specific: for instance, the technology sector is affected by industry-wide regulations and changes in customer demands for a given product.
Macroeconomic factors like interest rates or inflation may also impact entire sectors, causing the underlying stocks to react similarly.
Furthermore, exogenous news and events can impact stocks belonging to different economic fields.
We underline that even stocks that do not have fundamental similarities may show relevant correlation, possibly due to investor sentiment~\cite{baker2007investor,hirshleifer2003herd}.
During periods of market uncertainty or volatility, investors tend to adopt a risk-aware approach, affecting multiple stocks simultaneously.
Finally, another source of correlation comes from composite financial instruments such as Exchange-Traded Funds (ETFs).
Since these involve a basket of stocks, correlation among otherwise unrelated assets could arise as they align with the funds' overall performance~\cite{leippold2016index}.
%However, it is important to note that correlation does not imply causation. 
%While two stocks may exhibit a high correlation, it doesn't necessarily mean one stock directly affects the other.

To conclude, the existence of the described correlation dynamics cannot be neglected when designing AI tools for the generation of synthetic datasets to simulate financial markets. 
However, most of the existing approaches attempt to learn stock correlation by solely relying on specific deep learning model architectures (e.g., a transformer layer designed to learn feature dependency~\cite{tashiro2021csdi}). 
Yet, our experiments demonstrate that the resulting \ts do not always preserve the existing correlations. 

\begin{figure}[t]
    \centering
    \includegraphics[width=\linewidth]{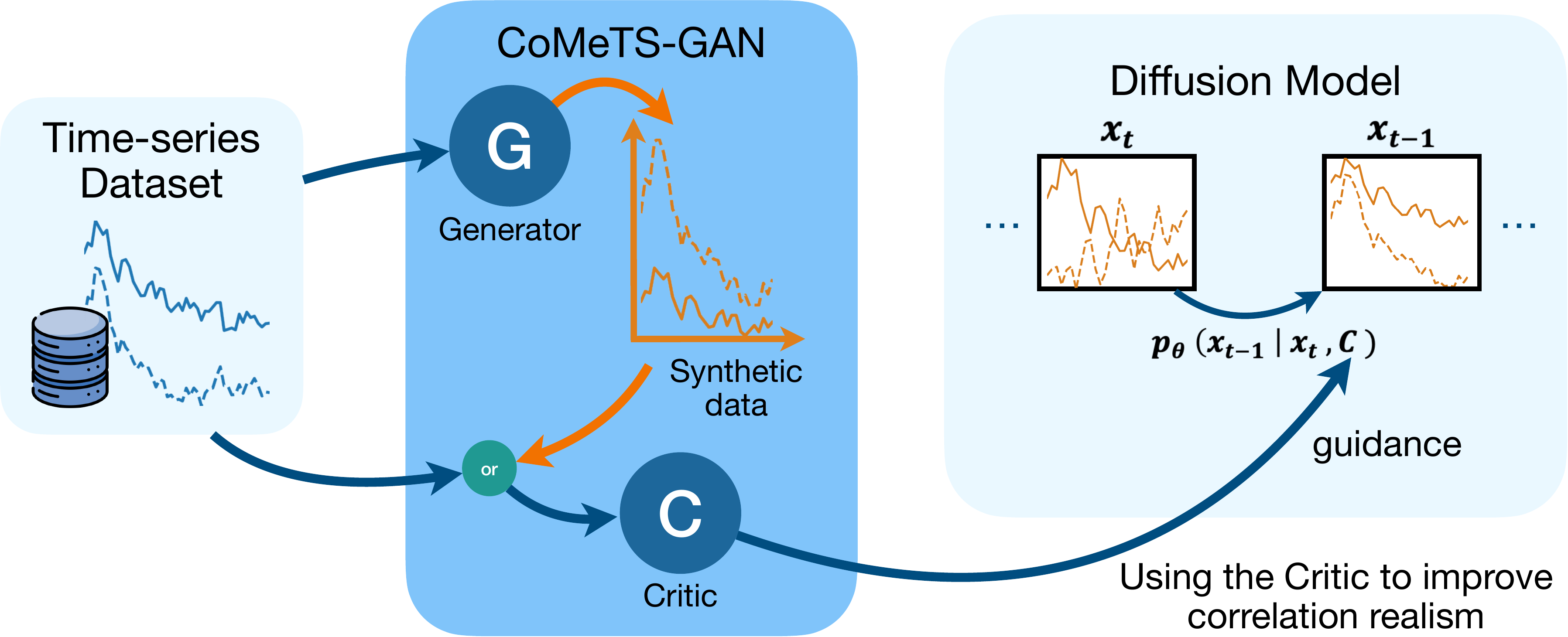}
    \caption{%Draft of the framework.
    Overview of the \framName framework. The system employs a conditional GAN consisting of a Generator (G) and a Critic (C).
    The Critic not only evaluates realism to properly train the Generator but also can guide a Diffusion Model during sample generation to improve the overall time-series quality.}
    \label{fig:all_arch}
    \vspace{-.3cm}
\end{figure}

\paragraph{\textbf{A novel quality-aware framework for correlated time-series generation}}
In this paper, we present a quality-aware generative framework that consists of a Generative Adversarial Network (GAN)~\cite{goodfellowGenerativeAdversarialNets2014} and a Diffusion Model~\cite{song2020score}.

Specifically,  we first introduce \framName (\textbf{Co}rrelated \textbf{M}ultivariat\textbf{e} \textbf{T}ime \textbf{S}eries \textbf{GAN}) for the synthetic generation of realistic market data (i.e., prices and volumes) that explicitly aims at providing a responsive market environment whilst capturing the inherent correlation among assets. Such approach builds on the GAN architecture~\cite{goodfellowGenerativeAdversarialNets2014}, which has been widely adopted for financial scenarios~\cite{vuletic2024volgan,cont2022tail,wieseQuantGANsDeep2020,takahashi2019modeling}, as well as in general-purpose \ts generation~\cite{seyfi2022generating,jeon2022gt,li2022tts}. Our proposed GAN architecture offers lightweight training and rapid generation, while also leveraging the two-player setup --- a \textit{Generator} and a \textit{Critic} --- to introduce domain-specific constraints. In particular, we incorporate cross-asset correlation coefficients directly into the Critic model, enabling the GAN to learn and reproduce realistic correlation dynamics. 

Eventually, we show that our Critic can be repurposed as an evaluation module --- considering its capability of assessing the realism of correlation dynamics in the generated time series --- and we demonstrate how it can be seamlessly integrated into a Diffusion-based Model for \ts generation to guide the sampling process at inference time. We show that such an integrated approach provides a quality-aware generation pipeline that enforces realistic correlation structures.
Figure~\ref{fig:all_arch} shows the overall proposed framework. 

\paragraph{\textbf{Paper Contributions}} In detail, the major contributions of this paper are the following:
\begin{itemize}
    \item A Quality-Aware generative framework consisting of a Conditional Generative Adversarial Network (CGAN) and a Diffusion Model for multivariate \ts generation, comprising both price and volume dynamics across multiple correlated assets.    
    
    \item A Conditional GAN architecture that uses a novel \textit{cross-correlation distance} to quantify the realism of correlation dynamics in generated \ts. A lightweight GAN architecture that enables fast training and generation, allowing for the synthesis of arbitrarily long \ts and responsive market simulation.

    \item A \textit{Critic-Guided} generation strategy for diffusion models, in which the GAN's Critic module dynamically steers the denoising process --- generation process --- of (any) pre-trained diffusion models at inference time, enhancing cross-correlation dynamics in the generated \ts.

    \item A thorough experimental campaign that points out the realism of our approaches, capturing all the known stylized facts and the correlation dynamics.  A comparison with existing approaches reveals that our solution outperforms existing work in its ability to capture inter-asset correlation, producing longer \ts, with moderate training effort.
    
    \item A simulation of counterfactual scenarios, enabling the generation of diverse \textit{what-if} financial \ts under hypothetical shifts in correlation dynamics.  We also evaluate the model's responsiveness to the behavior of experimental agents. 

\end{itemize}

\section{Related Work}
\label{sec:related}
In this section, we review existing generative models, and related work, aimed at generating \ts data, with a particular focus on methods developed for financial and stock price datasets.

\paragraph{Time series generation}
Early generative models for \ts focused on domains outside finance. 
C-RNN-GAN~\cite{mogrenCRNNGANContinuousRecurrent2016} utilized LSTM networks in both the generator and discriminator to produce sequences of classical music, while RCGAN~\cite{estebanRealvaluedMedicalTime2017} adapted a similar framework for medical \ts, introducing a conditioning mechanism distinct from simple auto-regression. 
These models rely mainly on binary adversarial feedback, which may be insufficient to capture complex temporal dependencies prevalent in financial data.

A notable advancement was proposed by Yoon et al.~\cite{yoon2019time} with TimeGAN, a novel framework combining adversarial objectives with stepwise supervised losses. 
TimeGAN employs four components (embedding function, recovery function, sequence generator, and sequence discriminator), demonstrating superior performance in terms of discriminative and predictive metrics on benchmark datasets. 
However, despite its merits, TimeGAN generates only fixed-length mini-sequences and is associated with high training costs, both of which are problematic for financial applications that require efficient generation of arbitrarily long sequences.

Other recent contributions have further advanced the field. 
GT-GAN~\cite{jeon2022gt} introduced a general-purpose framework for \ts synthesis based on GANs, designed to handle a diverse range of sequential data by explicitly modeling temporal patterns and dependencies across various domains. 
COSCI-GAN~\cite{seyfi2022generating} proposed a novel approach for generating multivariate \ts by coordinating the latent sources underlying different variables, thereby enabling the synthesis of multivariate sequences with common-source dependencies. 
While these models expand generative capabilities beyond single-variate and fixed-length sequences, they have yet to address the specific challenges of creating realistic, arbitrarily long financial \ts with dynamic interdependencies among multiple assets.

\paragraph{Diffusion models for Time-Series generation}
Recently, a new class of generative models, namely diffusion models, has been adopted for \ts generation. Diffusion models were originally introduced in~\cite{sohl2015deep} and gained widespread popularity through the Denoising Diffusion Probabilistic Models (DDPM) proposed in~\cite{ho2020denoising}.  They have demonstrated superior performance in sample quality and diversity across modalities~\cite{dhariwal2021diffusion}, and recent work has begun extending their applicability to structured domains such as \ts and dynamical data~\cite{lin2024diffusion}.
Notable examples include DiffTime~\cite{coletta2023constrained}, Diffusion-TS~\cite{yuan2024diffusion}, and  TSGM~\cite{lim2023regular} for the generation of \ts data.

\paragraph{Financial-specific generative models}
The aforementioned work mainly focuses on general-purpose generative models, which only partially adapt to the specific requirements of financial \ts generation. In fact, existing models often fail to capture all the stylized facts and statistical properties of financial data. 
Consequently, recent literature has introduced ad-hoc models specifically tailored to financial data.

Takahashi et al.\cite{takahashi2019modeling} proposed FIN-GAN, adopting classic GANs with a variety of generator/discriminator structures (multi-layer perceptrons, convolutional neural networks, and hybrids) to reproduce stylized facts observed in actual market \ts. 
Wiese et al.\cite{wieseQuantGANsDeep2020} developed QuantGAN, integrating temporal convolutional networks (TCNs) to effectively model long-range dependencies such as volatility clustering. 
Li et al.~\cite{li2022tts} explored transformer-based models for adversarial generation of time series, leveraging self-attention to improve sequence modeling. 
Nonetheless, none of these methods provides a scalable solution for generating multi-stock series that capture evolving interactions and dependencies among different assets.

Broadening the view, some works focused on the generation of specific aspects.
Vuletić et al.~\cite{vuletic2023fin} exploit a GAN architecture trained with an economic-driven loss function for the generator to generate the ETF-excess returns with respect to the underlying stocks to place trades and make profits.
Cont et al.~\cite{cont2022tail} proposed an approach based on GANs to generate price traces that retain tail risk features observed in the input dataset, to improve the estimation of loss distributions in dynamic portfolios.
However, these approaches do not consider the simultaneous generation of multiple stocks and their correlation dynamics.

Masi et al.~\cite{masi2023correlated} is the first work that explicitly addresses the generation of correlated stock prices, using a conditional GAN architecture. 
However, in their framework, volumes and prices are generated independently, which is suboptimal, and the intraday structure of trading activity is largely overlooked. 
Our quality-aware generation framework builds upon and extends this prior work in several aspects. 
First, we introduce novel temporal features to better capture intra-day stock dynamics (e.g., the characteristic U-shaped volume profile~\cite{bouchaud2018trades}). 
We extend the GAN framework to jointly generate trading volumes and prices, thereby enhancing realism by capturing the strong dependence between volume dynamics and price volatility. 
Our analysis of stylized facts is accordingly expanded to encompass empirical regularities in both volume and price series.
Finally, we went beyond a single model by also introducing diffusion models: we show that our trained Critic can be integrated into any existing diffusion model to guide the generation of \ts towards more realistic inter-asset correlations. 
Within this new framework, we also demonstrate the possibility of counterfactual generation, assessing the model’s ability to simulate previously unseen market scenarios - for instance, exploring a hypothetical shift in the correlation between Coca-Cola and Pepsi stock returns from positive to negative.

\section{Background}
\label{sec:background}
In this section, we briefly review some important pillars that are useful for the reader to approach this paper.

\subsection{Conditional GANs}
Generative Adversarial Networks (GANs)~\cite{goodfellowGenerativeAdversarialNets2014} are a subclass of generative models based on game theory.
GANs have two neural networks: a generator denoted as $G$, and a discriminator denoted as $D$.
The generator creates synthetic data samples from random noise, while the discriminator distinguishes between real and generated samples.
During training, the generator competes against the discriminator to improve its ability to generate synthetic data samples $\bm{x} \sim p_\text{gen}$ that are realistic, i.e., distributed as the original data distribution $p_\text{real}$. 

The generator's output is $\bm{x} = G(\bm{z}; \bm{\theta}^{(G)})$ where $\bm{z} \sim \mathcal{\bm{N}}(\bm{0}, \bm{I})$ is an input noise to ensure variance at generation time, and $\bm{\theta}^{(G)}$ the parameters of the generator. 
The discriminator outputs $y = D(\bm{x}; \bm{\theta}^{(D)}) \in [0, 1]$ representing the probability of $\bm{x}$ being drawn from $p_\text{real}$ and $\bm{\theta}^{(D)}$ the parameters of the discriminator.
Both $G$ and $D$ are trained to optimize their payoff until neither player can unilaterally improve its cost.
The discriminator is trained to maximize the probability of assigning the correct label to the data, either \textit{real} or \textit{generated}; conversely, the generator is trained to minimize it (by generating realistic samples that the discriminator misclassifies as real).

GANs can also be extended to consider conditional input, resulting in what is known as conditional-GANs~\cite{mirzaConditionalGenerativeAdversarial2014}.
In conditional-GANs, or CGANs, both the generator and discriminator networks may take additional conditioning information $\bm{c}$ as input, such as class labels or other contextual features.
This allows for generating specific types of data samples based on the provided conditioning information.
The generator's and discriminator's input vectors are combined with $\bm{c}$.

In this work, we use a Wasserstein Generative Adversarial Network (WGAN)~\cite{arjovskyWassersteinGenerativeAdversarial2017}, which demonstrated better training performance and stability. 
In a WGAN the discriminator is called critic and outputs a realness score of the input samples, $y = D(\bm{x}; \bm{\theta}^{(D)}) \in \mathbb{R}$, even called \textit{critic score}.
Formally, the optimal generator is $G^*$:
%\vspace{-.2cm}
\begin{equation}
    \arg \min_G \max_D \quad \mathop{\mathbb{E}}_{\bm{x} \sim p_{\text{data}}} [ D(\bm{x} | \bm{c})] - \mathop{\mathbb{E}}_{\bm{z} \sim \mathcal{N}(\bm{0}, \bm{I})} [D(G(\bm{z} | \bm{c}))].
    \label{eq:wgan_loss}
\end{equation}

\subsection{Diffusion Model}
% pass \cite{song2020score, dhariwal2021diffusion}.
Diffusion models generate data by learning to invert a Markovian noising process that maps data to a tractable prior, such as a standard Gaussian.

Let $\bm{x} \sim p_{\text{real}}$ be a real sample. 
The \emph{forward diffusion process} progressively corrupts $\bm{x}_0$ into a sequence of latent variables $(\bm{x}_1, \ldots, \bm{x}_T)$ using a Markov chain: $q(\bm{x}_{1:T}|\bm{x}_0) = \prod_{t=1}^T q(\bm{x}_t|\bm{x}_{t-1})$.
Each transition is typically defined as additive Gaussian noise: $ q(\bm{x}_t|\bm{x}_{t-1}) = \mathcal{N}\left(\bm{x}_t; \sqrt{1-\beta_t}\bm{x}_{t-1}, \beta_t \bm{I}\right)$, with a time-dependent noise schedule $\{\beta_t\}_{t=1}^T$.
After $T$ steps, $\bm{x}_T$ becomes approximately standard Gaussian, i.e., $q(\bm{x}_T) \approx \mathcal{N}(\bm{0}, \bm{I})$ for large $T$.

The \emph{reverse process} aims to reconstruct data from noise by learning the reverse conditional distributions:
$p_\phi(\bm{x}_{0:T}) = p(\bm{x}_T) \prod_{t=1}^T p_\phi(\bm{x}_{t-1}|\bm{x}_t)$, where $p(\bm{x}_T) = \mathcal{N}(\bm{0}, \bm{I})$ and each $p_\phi(\bm{x}_{t-1}|\bm{x}_t)$ is typically parameterized as
$p_\phi(\bm{x}_{t-1}|\bm{x}_t) = \mathcal{N}\left(\bm{x}_{t-1}; \bm{\mu}_\phi(\bm{x}_t, t), \bm{\Sigma}_\phi(\bm{x}_t, t)\right)$, with the mean and covariance learned via a neural network $\epsilon_{\phi}$.

In practice, this neural network is trained by maximizing a variational lower bound on the marginal log-likelihood $\log p_\phi(\bm{x}_0)$, yielding a loss that includes a (reweighted) denoising score-matching term~\cite{ho2020denoising}:
%\vspace{-.2cm}
$$\mathbb{E}_{t, \bm{x}_0, \epsilon} \left[ \left\| \epsilon - \epsilon_\phi \left(\bm{x}_t, t \right) \right\|^2 \right],
%\vspace{-.2cm}
$$
where $\epsilon \sim \mathcal{N}(\bm{0}, \bm{I})$ and $\bm{x}_t = \sqrt{\bar{\alpha}_t} \bm{x}_0 + \sqrt{1 - \bar{\alpha}_t} \epsilon$, with $\bar{\alpha}_t = \prod_{s=1}^t (1-\beta_s)$.

Sampling proceeds by initializing $\bm{x}_T \sim \mathcal{N}(\bm{0}, \bm{I})$ and sequentially applying the learned reverse transitions $p_\phi(\bm{x}_{t-1}|\bm{x}_t)$ for $t = T, \ldots, 1$, reconstructing data via iterative denoising.

\subsection{Stylized Facts}
\label{subsec:stylized_facts}
The variation of asset price exhibits several statistical properties that are common across a wide range of markets and timeframes. 
These properties capture the market behavior over different time periods and are referred to as \textit{stylized facts}~\cite{cont2001empirical, vyetrenko2020get}.
In this work, we use them to evaluate the quality of samples generated by the model. 
In particular, we consider the following stylized facts:

\begin{itemize}
    \item \textbf{Fat-tailed distribution} or \textbf{heavy tails}.
    The distribution of the asset returns displays a power-law or Pareto-like tail, meaning that extreme returns occur more frequently than predicted by a normal distribution.
    
    \item \textbf{Aggregational normality}.
    As the time period for computing returns ($\Delta t$) increases, the distribution of returns resembles a normal distribution.
    \item \textbf{Linear unpredictability} or \textbf{absence of autocorrelation}. 
    The linear autocorrelation of the return $\mathrm{corr}(r_{t, T}, r_{t + \Delta t, T})$ rapidly decays to zero within minutes and becomes insignificant for periods longer than $\Delta t = 20$ minutes, with $T$ the length of the series~\cite{vyetrenko2020get}.
    \item \textbf{Volatility clustering}. 
    Large changes in prices tend to cluster together, that is, volatility shows a positive autocorrelation over several trading days.
    \item \textbf{Volume/Volatility Correlation}.
    Trading volume and volatility are positively correlated.
\end{itemize}
Specifically, these properties allow us to assess the preservation of the realism of generated traces. 

\subsection{Correlation Dynamics}
Interdependency and correlation dynamics need new analytical tools to analyze the set of stocks in relation to each other. We recall the importance of correlation dynamics in portfolio management, the use of correlation as a measure to evaluate portfolio diversification is a common practice. The principal aim is to mitigate the total risk by strategically including assets that exhibit a low correlation with each other. To achieve this, it is imperative to maintain a realistic understanding of the correlation dynamics among stocks.

In order to evaluate the effectiveness of the model in capturing these dynamics, we suggest the following quantitative metric, the {\em cross-correlation distance}: % described in Section~\ref{par:ccd}.

\textbf{Cross-correlation distance.}
\label{par:ccd}
This metric measures the distance between the cross-correlations of real and generated stocks. 
Formally, let $\rho(\cdot, \cdot)$ be the \textit{Pearson correlation coefficient} (\textit{PCC}).
For a pair of stocks $(S_i, S_j)$, we can measure:
%\vspace{-.2cm}
$$d_{\rho}(S_i, S_j) = \mathrm{MSE} (\rho(S_i, S_j), (\rho(\widehat{S_i}, \widehat{S_j})),
%\vspace{-.2cm}
$$
where $\widehat{S_i}$ and $\widehat{S_j}$ are the synthetic version of the stocks $S_i$ and $S_j$ and MSE being the mean squared error. 

We will utilize this metric to scrutinize the model's performance with respect to stocks that show varying degrees of correlation.
For instance, let us consider two prominent companies, Coca-Cola and PepsiCo, which generally demonstrate a strong positive correlation. 
This correlation is attributable to their simultaneous presence in the same market, competing for similar consumer demographics, and facing exposure to similar exogenous factors (i.e., suppliers).
On the other hand, during times of market turmoil, investors often seek out defensive stocks as opposed to riskier, growth-oriented stocks.
This pattern of trading behavior may result in a shift to Coca-Cola, a beverage firm, from a technology company, for instance, Nvidia, leading to a negative correlation between these two stocks.

Understanding these correlation patterns is not just academically interesting, it is critical for managing risk and unlocking profit opportunities.

\section{System Model}
\label{sec:problem_formulation}
The purpose of our framework is to generate multivariate \ts capturing crucial aspects of the stock markets for multiple stocks simultaneously, which include mid-prices and volumes.
A key focus of our approach is to ensure the preservation of the correlation between the considered stocks. 
The framework's underlying model is trained to maximize both the similarity and the match in the correlation of the generated \ts to real ones.
This architecture can be used to generate \ts also of other domains. 
Nonetheless, we will mostly stress its strengths in the financial field; more specifically, we are interested in mid-prices and volumes of a chosen set of stock.
In the following, we analyze in detail the main components of the framework.

\paragraph{Output} 
The output of our framework is a multivariate \ts of length $F$ units of time, two series (price and volume) for each stock in the set $\mathcal{S}$. 
This output is structured as a matrix \mbox{$\bm{\hat{x}}_{\text{future}} \in \mathbb{R}^{F \times 2n}$}, with $n=|\mathcal{S}|$ (i.e., the number of stocks). 
To generate longer \ts, an auto-regressive approach can be used, that is, iteratively feeding the conditional generator with its own output. %{\color{red} Nell'architettura richiediamo che l'input sia la finestra precedente? è quindi sempre "conditional" generation? What if I want to generate a new time-series from just noise? E poi come facciamo questo su CDSI? facciamo cond. generation as well direi, non auto-regressive?}
More details on the auto-regressive generation are provided in the following paragraphs.

\paragraph{Training Data}
%We will elaborate on the dataset generation process specifically for the mid-price \ts. 
%However, the methodology is analogous for volumes as well.
The dataset used to train the model is generated from historical stock traces in the temporal period $[0, T]$. 
Let $\boldsymbol{x}_s \in \mathbb{R}^{2 \times T}$ be the \ts representing the mid-price and the volume of stock $s \in \mathcal{S}$ in the considered temporal period.
For the given set of stocks, we stack the individual \ts $\boldsymbol{x}_s$ along the columns to create the multivariate \ts $\mathbf{X}_\mathcal{S} \in \mathbb{R}^{T \times 2n}$.
In order to capture the dependencies between past and future stock movements, we segment the multivariate \ts $\mathbf{X}_\mathcal{S}$ over time to construct the dataset as follows.
For each time-step $t$, we create a pair consisting of the past $P$ time-steps, denoted as $\bm{x}_{\mathrm{past}}$ and the future $F$ steps denoted as $\bm{x}_{\mathrm{future}}$.
By applying this segmentation for $t \in [P, T-F]$, we generate the dataset which we denote as:
\begin{equation}
    \mathcal{D} = \{ (\mathbf{X}_\mathcal{S}[t-P:t], \;   \mathbf{X}_\mathcal{S}[t+1: t+1+F] )\}_{t=P}^{T-F}
\end{equation}
The objective of the model is to generate a continuation $\bm{\hat{x}}_{\text{future}}$ of an observed \ts $\bm{x}_{\mathrm{past}}$, by training the model on tuples of the form $(\bm{x}_{\mathrm{past}}, \bm{x}_{\mathrm{future}}) \in \mathcal{D}$.
The simultaneous generation of $2n$ \ts allows the preservation of the correlation dynamics observed in the training set. 

\begin{figure}
    \centering
    \includegraphics[scale=.5]{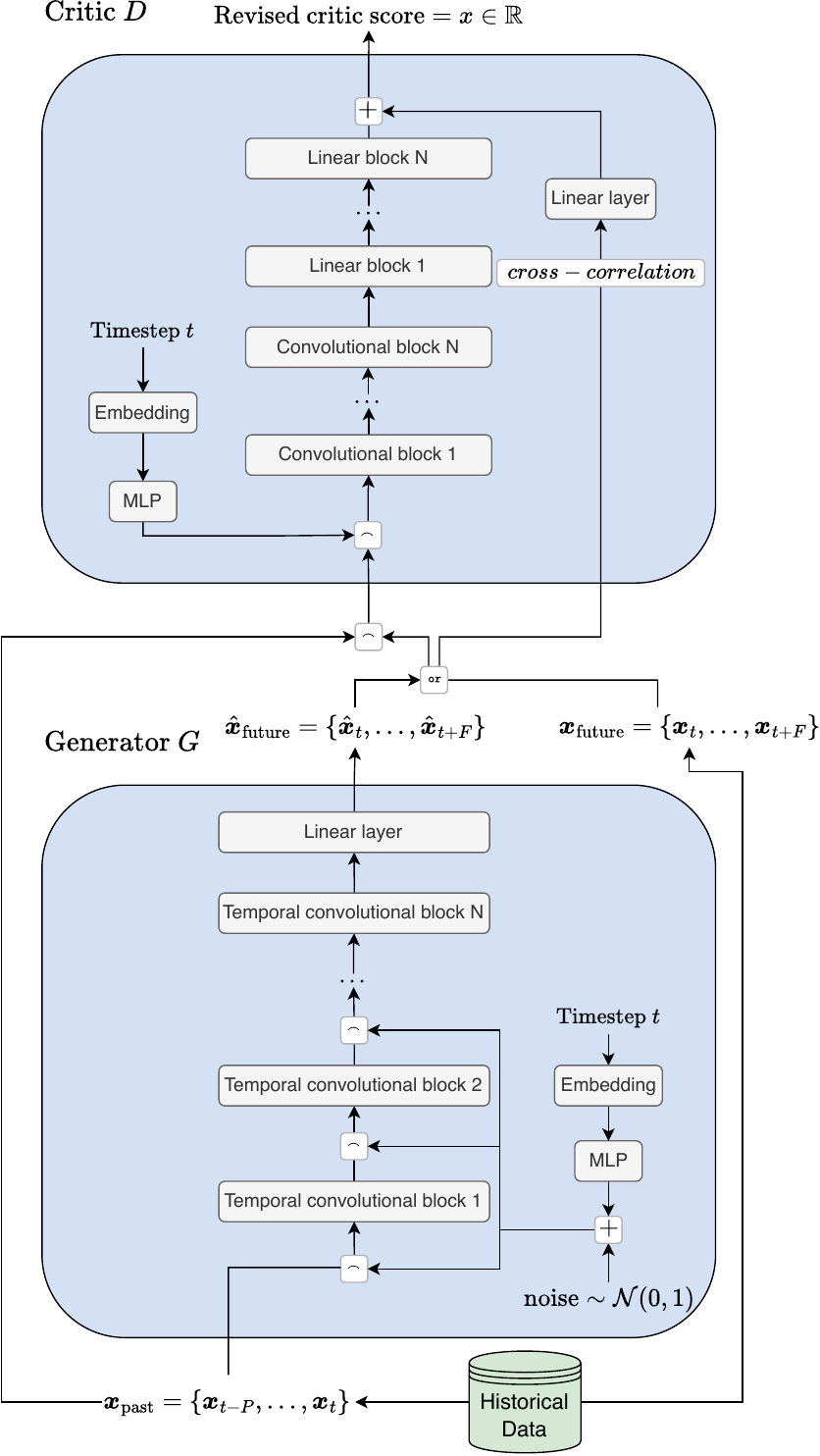}
    \caption{C-WGAN architecture. The $\frown$ operator represents the concatenation of two vectors. The \texttt{or} operator iteratively alternates between the real series $\bm{x}_\mathrm{future}$ and the generated series $\hat{\bm{x}}_\mathrm{future}$.}
    \label{fig:gan_architecture}
\end{figure}

\subsection{\framName Architecture}
The underlying model of \framName is a Conditional Wasserstein Generative Adversarial Network (C-WGAN)~\cite{mirzaConditionalGenerativeAdversarial2014, arjovskyWassersteinGenerativeAdversarial2017}.
This model is trained to learn the joint probability distribution of the collection of future \ts $\bm{x}_{\mathrm{future}}$ in the training dataset $\mathcal{D}$ by optimizing the critic score.
%{\color{red}We adopt spectral normalization to enforce the Lipschitz continuity constraint, due to its superiority to other techniques like weight clipping or gradient penalty and for its ease of applicability. }
%\gm{In realtà ora usiamo anche gradient penalty, seguendo l'algoritmo in~\cite{gulrajani2017improved} oltre a spectral normalization~\cite{miyato2018spectral}.}
As for the generator and the critic of the C-WGAN, we used a mixture of vanilla and temporal convolutional neural networks.
Figure~\ref{fig:gan_architecture} shows the architecture of our model. 

The \textit{\textbf{generator}} $G$ is the main component of \framName.
It is responsible for the actual generation of the \ts $\bm{\hat{x}}_{\text{future}}$, which retains the statistical properties of the real \ts observed by the generator.
The architecture used for the generator network is composed of a series of 7 temporal convolutional blocks~\cite{bai2018empirical} with increasing dilation size and a final linear layer to adjust the size of the output vector. 
Each temporal block is followed by a \texttt{leaky ReLU} activation function and a \texttt{dropout} layer.

The \textit{\textbf{critic}} $D$ is responsible for assessing the realness of the generated samples. 
The architecture used for the critic network is made of a series of convolutional layers having increasing filter sizes, followed by a series of linear layers. 
Each convolutional and linear layer is followed by a \texttt{leaky ReLU} activation and a \texttt{dropout} layer, and spectral normalization is applied on top of each network layer. 
No activation function is used for the last layer of the network.
We denote the output of this sequential network as $o_1$, which is devoted to measuring the realness of the generated \ts.
A linear layer is present within the critic (top right of the figure) and is meant to address the correlations of the multivariate \ts.
The linear layer takes as input the correlation coefficients 
%(even referred to as \textit{cross-correlation} values) 
computed between all pairs of the mid-prices and volumes of the $n$ stocks (i.e., $\binom{2n}{2}$ pairs) and it outputs $o_2$. 

The overall output of the critic (i.e., the critic score) is the sum $o = o_1 + \alpha \cdot o_2$, with $\alpha \in [0,1]$ being a hyperparameter to weigh the relevance of the correlation.
By summing these two scores, we aim to capture the realism and correlation of the $n$ stocks.

\paragraph{\framName Training.}
Upon input pair 
$(\bm{x}_{\mathrm{past}}, \bm{x}_{\mathrm{future}}) \in \mathcal{D}$, the framework is trained as follows.
The input of the generator is the concatenation of the past sequence $\bm{x}_\mathrm{past}$ and a noise vector $\bm{z} \sim \mathcal{N}(\bm{0}, \bm{I})$ of the same length summed to a vector $\bm{t}$ to embed the time information.
The temporal information is computed by first mapping each timestamp to the number of minutes elapsed since the start of the trading day. 
This value is then discretized into uniform intervals, each representing a fixed duration (10 minutes), effectively segmenting the trading session into a sequence of time bins. 
This variable is then mapped to a high-dimensional embedding using a sinusoidal positional encoding scheme, similar to those used in transformer architectures.
In the generator, the time-step embedding is processed by a 2-layer MLP with a non-linear activation function ($f(x)=x\cdot \frac{e^x}{1+e^x}$) to further enhance its expressiveness. 
The resulting time embedding is added directly to the noise, effectively conditioning the generative process on the temporal context.
This integration allows the model to generate outputs that are sensitive to the progression of time within each sequence, improving its ability to model temporal dynamics.

Each temporal block is fed with the output of the previous, concatenated with the noise vector $z$ along the dimension of the features. 
The noise injection technique allows diversity in the model's output.
Finally, the output of the last temporal block is directly fed to the last linear layer. 
The output of the generator network is the future sequence $\hat{\bm{x}}_\mathrm{future}$.

The input of the critic is the concatenation of the past sequence $\bm{x}_\mathrm{past}$ with either the real continuation $\bm{x}_\mathrm{future}$ drawn from the dataset, or a generated one $\hat{\bm{x}}_\mathrm{future}$ from the generator. 
In the discriminator, the temporal information is processed in the same way of the generator, but the resulting time embedding is concatenated as an additional channel to the input sequence along the feature dimension.
%This approach allows the network to access explicit temporal context at every layer, enabling it to better capture and utilize the temporal structure present in the data during discrimination.

The result of the concatenation is fed to the convolution and linear blocks to produce a single value in output, representing the standard critic score $o_1$.
Additionally, the correlations between the features of the future sequence in the input are computed.
All the $\binom{2n}{2}$ correlation coefficients are input to the linear layer to produce the score $o_2$. 
The output of the critic is the sum of the two contributions, as mentioned above.

We adopt spectral normalization to enforce the Lipschitz continuity constraint, due to its superiority to other techniques like weight clipping or gradient penalty and for its ease of applicability~\cite{miyato2018spectral}.

\subsection{Diffusion Models: A \textit{Critic-Guided} generation}
Sohl-Dickstein et al.~\cite{sohl2015deep} and Song et al.~\cite{song2020score} show that any pre-trained diffusion model can be conditioned using the gradients of a classifier, steering the denoising (i.e., generation) process towards an arbitrary class label y. This technique, called classifier guidance, in diffusion models typically involves modifying the score function to incorporate the gradient from an external classifier that estimates the likelihood that a sample belongs to a target class. 

In \framName, we note that the Critic module is trained to distinguish between real and synthetic stock price sequences; it implicitly learns a representation of realism. Therefore, we propose to utilize our Critic as a surrogate classifier, where its gradient can be leveraged to enhance the generation process of the diffusion model. Mathematically, the diffusion model’s score function, denoted as $s(x_t, t)$, is adjusted by incorporating the gradient of the WGAN critic $D(x)$: 
%\vspace{-.2cm}
\begin{equation}
\tilde{s} = s(x_t,t)+w\cdot \nabla_{x_t}D(x_t),
\label{eq:guidance}
%\vspace{-.2cm}
\end{equation}
where $w$ modulates the guidance strength. 
This modification encourages the diffusion process to produce samples that align more closely with the distribution learned by the WGAN critic, effectively biasing the sampling trajectory toward regions of higher realism.

In the experimental section, we instantiate our \textit{Critic-Guided} strategy using DiffTime~\cite{coletta2023constrained}, a state-of-the-art Diffusion Model trained on our \ts datasets. We demonstrate how the gradients of the Critic can actually guide the diffusion denoising process at inference time, improving the cross-correlation fidelity of the generated \ts and yielding a quality-aware generation pipeline. 
More broadly, our approach suggests that any differentiable generative network trained on a domain-specific realism metric can be used to enhance the generation quality of general-purpose diffusion models.

\section{Experiments}
\label{sec:experiments}
In this section, we evaluate the performance of our framework against SOTA approaches.
We first compare \framName using several benchmarks and newly curated datasets to ensure a rigorous assessment. Our primary focus is assessing the fidelity of the generated \ts in comparison to real ones. Eventually, we examine the improvement in generation realism achieved by the \textit{Critic-Guided} strategy, instantiated using DiffTime~\cite{coletta2023constrained} a recent diffusion models for time-series generation. 
We ran our Python-based framework on a NVIDIA 2060 GPU.\footnote{We commit to publicly releasing our source code upon acceptance, enabling other researchers to reproduce our results and use the framework for their own experiments.}

%  available on GitHub\footnote{GitHub repository \url{https://github.com/giuseppemasi99/COMETS-GAN}.}, 

\subsection{Datasets}
We test the performance of our framework across data showing varying periodicity, discreteness, level of noise, regularity of time-steps, and correlation across time and features. 
We also include real stock market \ts.
In detail, the considered datasets --- including well-known benchmarks from the literature --- are the following:

\begin{itemize}

    \item \textit{Sines}~\cite{yoon2019time}:
    \label{par:dataset_sines}
    Multivariate sinusoidal sequences with different frequencies $\eta$ and phases $\phi$. 
    Specifically, the dataset is obtained with \mbox{$s_i (t) = \sin (2\pi \eta_i t + \theta_i)$}, where $i \in \{ 1, \dots, 5 \}$, $\eta_i \sim \mathcal{U}[0, 1]$ and $\theta_i \sim \mathcal{U}[- \pi, \pi]$.

    \item \textit{Multivariate Gaussian Model}~\cite{yoon2019time}:
    \label{par:dataset_gaussian}
    Sequences from auto-regressive multivariate Gaussian models, defined as: $g_i(t) = \phi_i g_i(t - 1) + q$ where $q \sim \mathcal{N}(0, \sigma_i + (1 - \sigma_i))$, $\phi_i \in [0, 1]$, and $\sigma_i \in [-1, 1]$.
    The coefficients $\phi_i$ and $\sigma_i$ allow to control the correlation across time and features, respectively.

    \item \textit{Stock Mid-Prices \& Volumes}~\cite{lobster}:
    \label{par:dataset_multistock} 
    The source of stock market data is LOBSTER\footnote{LOBSTER \url{https://lobsterdata.com/}}, an online limit order book data tool to provide limit order book data for the NASDAQ traded stocks.
    From limit order book data, it is easy to compute the mid-price with the temporal resolution of choice, that in our case, is minute.
    We focus on four stocks, namely, Coca-Cola (KO), PepsiCo (PEP), Nvidia (NVDA), and Kansas City Southern (KSU).
    The time period of choice is from 2018-02-02 to 2018-10-11 for the training set, and from 2018-10-12 to 2018-11-14 for the validation set.
    Table~\ref{table:stock_correlation} shows the Pearson correlation coefficient between them. 

    We observe how some are highly positively correlated (KO and PEP, NVDA and KSU), and some are highly negatively correlated (PEP and NVDA, PEP and KSU, KO and KSU, KO and NVDA).
    To test the scalability of our generative framework, we selected the 30 stocks composing the DJIA index from 2017-02-02 to 2017-10-11 for the training set and from 2017-10-12 to 2017-11-14 for the validation set.
    
    For these \ts, we follow the same preprocessing pipeline.
    For the mid-prices, we compute the log-returns normalized using a $z$-score approach.
    Instead, volumes are scaled in the $[-1, 1]$ with a \texttt{min-max} approach, then followed by \texttt{tanh} activation function, which scales the volumes to the positive domain.
\end{itemize}

\begin{figure}[h!]
\centering
\begin{tabular}{l|cccc}
\toprule
& KO & PEP & NVDA & KSU \\ 
\midrule
KO & $1.0$ & $0.94$ & $-0.66$ & $-0.52$ \\
PEP & $0.94$ & $1.0$ & $-0.81$ & $-0.56$ \\
NVDA & $-0.66$ & $-0.81$ & $1.0$   & $0.65$ \\
KSU & $-0.52$ & $-0.56$ & $0.65$  & $1.0$ \\
\bottomrule
\end{tabular}
\captionof{table}{Pearson correlation between the stock prices in the validation set of the multi-stock dataset.}
\label{table:stock_correlation}
\end{figure}

\begin{comment}
    
\begin{algorithm}[hbt]
    \caption{\textit{Guided-DiffTime}}
    \label{alg:guidingddim}
    \begin{algorithmic}
        \STATE Input: Critic $f_c : \mathcal{X} \rightarrow \mathbb{R}$, scale parameter $\rho$
        \STATE Output: new TS, $\mathbf{x}_0$
        \STATE $\mathbf{x}_T \gets \text{sample from } \mathcal{N}(0, \mathbf{I})$
        \FORALL{$t$ from $T$ to 1}
            \STATE $\hat\epsilon \gets \epsilon_{\theta}(\mathbf{x}_t, t)$
            \STATE $\hat\epsilon \gets \hat\epsilon - \rho \sqrt{1-\hat{\alpha}_t} \grad_{\mathbf{x}_t} f_c(\frac{1}{\sqrt{\hat{\alpha}_t}}(\mathbf{x}_t - \hat\epsilon \sqrt{ 1 - \hat{\alpha}_t}))$
            \STATE $\mathbf{x}_{t-1} \gets \sqrt{\hat{\alpha}_{t-1}} \left( \frac{\mathbf{x}_t - \sqrt{1-\hat{\alpha}_t} \hat{\epsilon}}{\sqrt{\hat{\alpha}_t}} \right) + \sqrt{1-\hat{\alpha}_{t-1}} \hat{\epsilon}$
        \ENDFOR
        \RETURN $\mathbf{x}_0$
    \end{algorithmic}
\end{algorithm}

\end{comment}

\subsection{Evaluation Metrics}
We measure the performance of our framework in terms of the following aspects: \textit{Realism}, \textit{Diversity}, \textit{Stylized Facts}, \textit{Reactivity}, \textit{Scalability}. 

\textbf{Realism.} We measure the realism of generated \ts by using the \textit{Discriminative Score}, which offers a quantitative measure of the similarity between \textit{real} and \textit{synthetic} (or \textit{generated}) \ts. 
It was proposed in~\cite{yoon2019time} and is computed as follows.
An LSTM is trained to differentiate between sequences from the original and generated datasets.
A standard supervised learning task is performed and the resulting classification error on the test set (the lower the better) provides a quantitative measure of similarity.

\textbf{Diversity.}
This aspect refers to the range and variability of patterns and structures present in the generated \ts data. 
It is crucial because a well-working generative model should produce data that fully explores the natural variations present in real \ts, rather than simply replicating or memorizing the training data (problem known as ``mode collapse").

\textbf{Cross-correlation distance.}
%\label{par:ccd}
In order to examine the model's ability to capture the correlation dynamics that exist between stocks we measure the distance between the cross-correlations of real and generated stocks using the cross-correlation distance defined in \ref{par:ccd}. 
Formally, at the end of each epoch, for each of the $\binom{2n}{2}$ pairs of features $(S_i, S_j)$, we measure the $d_{\rho}(S_i, S_j)$ with respect to their synthetic counterparts.

%$$d_{\rho}(S_i, S_j) = \mathrm{MSE} (\rho(S_i, S_j), (\rho(\widehat{S_i}, \widehat{S_j})),$$
%where $\widehat{S_i}$ and $\widehat{S_j}$ are the synthetic version of the stocks $S_i$ and $S_j$ and MSE being the mean squared error. 

\textbf{Stylized Facts.}
We investigate the properties of the synthetic traces by analyzing the most common stylized facts in the financial domain as discussed in Section~\ref{subsec:stylized_facts}.

\subsection{\framName Results}

\subsubsection{Realness of Generated Data}
We can generate arbitrarily long stock traces thanks to the \textit{auto-regressive application} of the model to the input \ts.
The generation process can be defined as follows: 
(1) the model is fed with a multivariate  of length $P$, that is a tensor in $\mathbb{R}^{P \times 2n}$ of original stock data; 
(2) the output of the model, is a multivariate \ts of length $F$ that is a tensor in $\mathbb{R}^{F \times 2n}$ of synthetic stock data; 
(3) the last $P$ time-steps of the synthetic sequence are fed to the model;
(4) steps 2 and 3 are repeated until a total number of generative steps are done.

\begin{figure}[h]
    \centering
    \includegraphics[width=.7\linewidth]{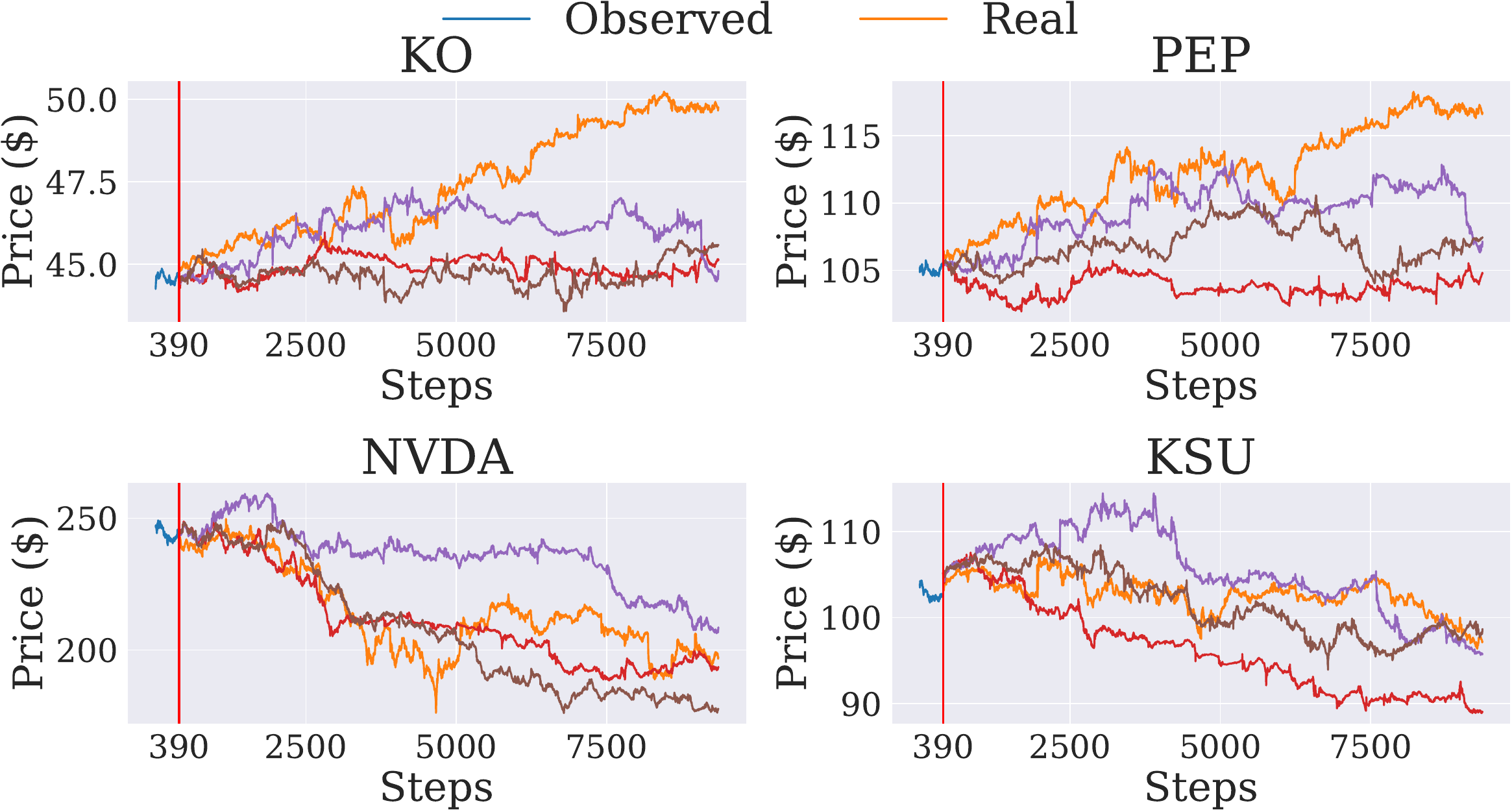}
    \caption{Diversity in price generation.}
    \label{fig:diversity_prices}
\end{figure}

\begin{figure}[h]
    \centering
    \includegraphics[width=.7\linewidth]{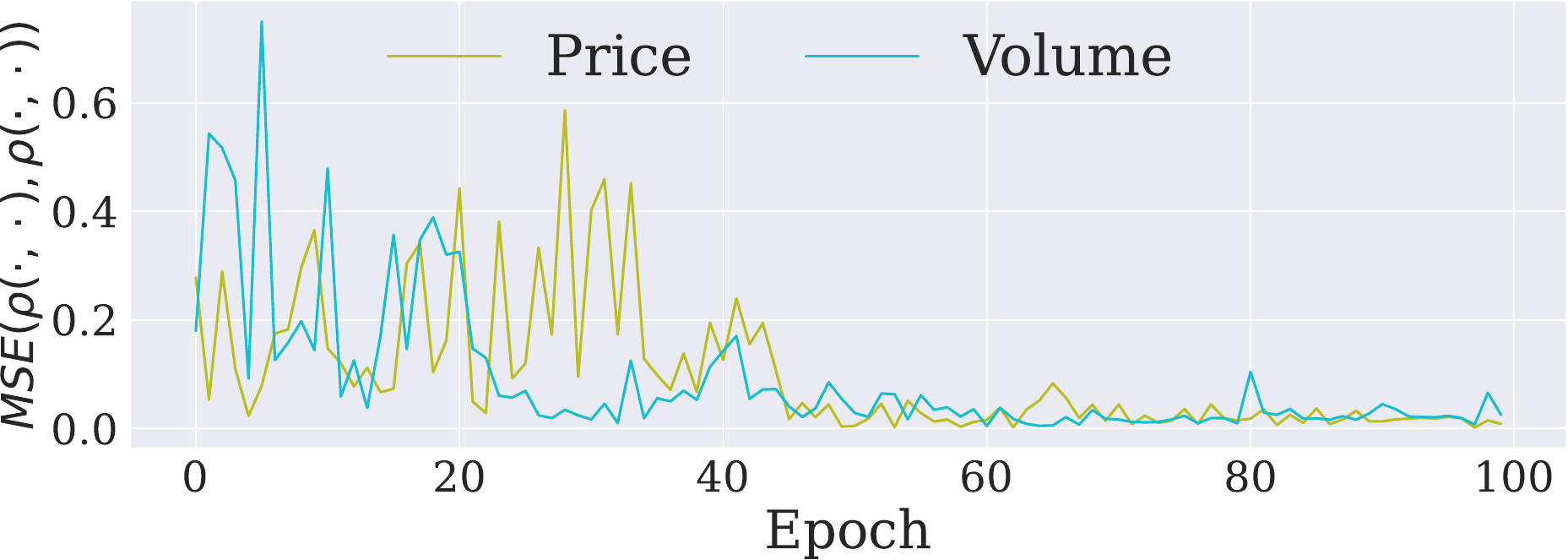}
    \caption{Average cross-correlation distance during training.}
    \label{fig:avg_corr_dist_val}
\end{figure}

By varying the random seed we are able to generate several multivariate \ts.
Figure~\ref{fig:diversity_prices} shows the execution of three runs, in addition to the real trace (orange line).
As we can observe, the model does not incur self-induction (i.e., it does not generate a trend that remains indefinitely influenced by it). 
Furthermore, the model respects the diversity property, generating different sequences from the same input if the random noise changes, preserving the correlation dynamics. %as will be further studied later on in this paper.
In fact, as the training of the model proceeds, the average cross-correlation distance between real and synthetic data decreases, as we can see in Figure~\ref{fig:avg_corr_dist_val}, becoming negligible among all the pairs of stocks. 
%\vspace{-.8cm}
%\vspace{-.3cm}

\begin{figure}
    \centering
    \includegraphics[width=.7\linewidth]{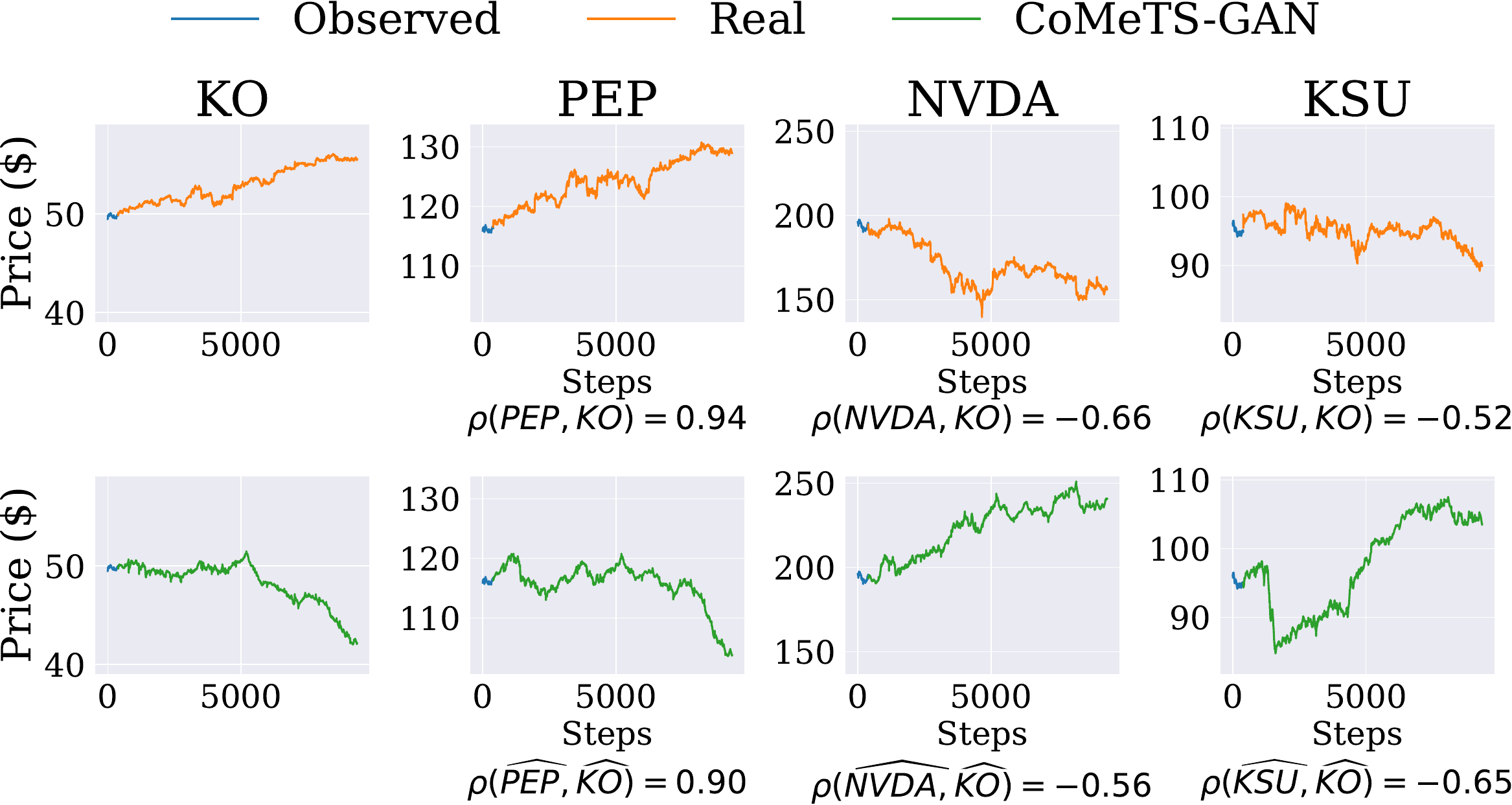}
    \caption{Price - Correlation between KO and the other stocks. While the generated data shows a downward trend, PEP and KO \ts keep the same correlation property of real data.}
    \label{fig:corr_wrt_KO}
\end{figure}

\begin{figure}
    \centering
    \includegraphics[width=.7\linewidth]{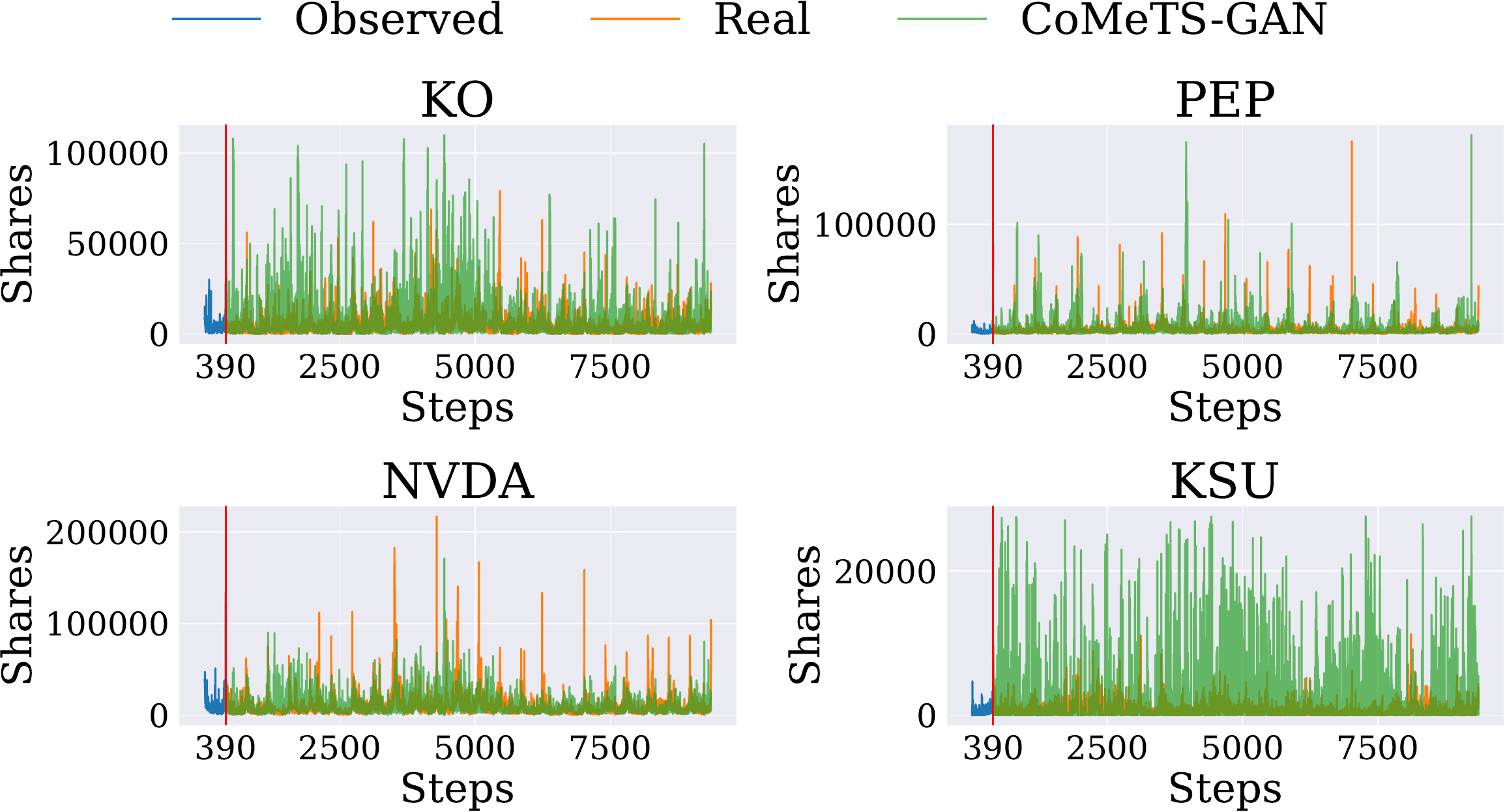}
    \caption{Real and Synthetic volumes - the synthetic data is able to partially reproduce the U-shaped volume pattern observed in the real data.}
    \label{fig:volumes}
\end{figure}

The fact that the model is able to capture the correlation dynamics is also evident in Figure~\ref{fig:corr_wrt_KO}, where we plot the prices of the stocks, and the average cross-correlation distance is $0.04$.
Figure~\ref{fig:volumes} shows the corresponding volumes.
%\clearpage

We measure the stylized facts on the auto-regressive generation of stock mid-prices and volumes for 24 trading days, given that a trading day lasts for 6.5 hours, we generate 9360 minutes in total. Notice that most of existing approaches are unable to generate such a long stock \ts data. Results are averaged over 10 runs where we vary the seed and the randomization of the generation. 

\begin{figure}
    \centering
    \begin{subfigure}{.49\linewidth}
        \centering
        \includegraphics[width=\linewidth]{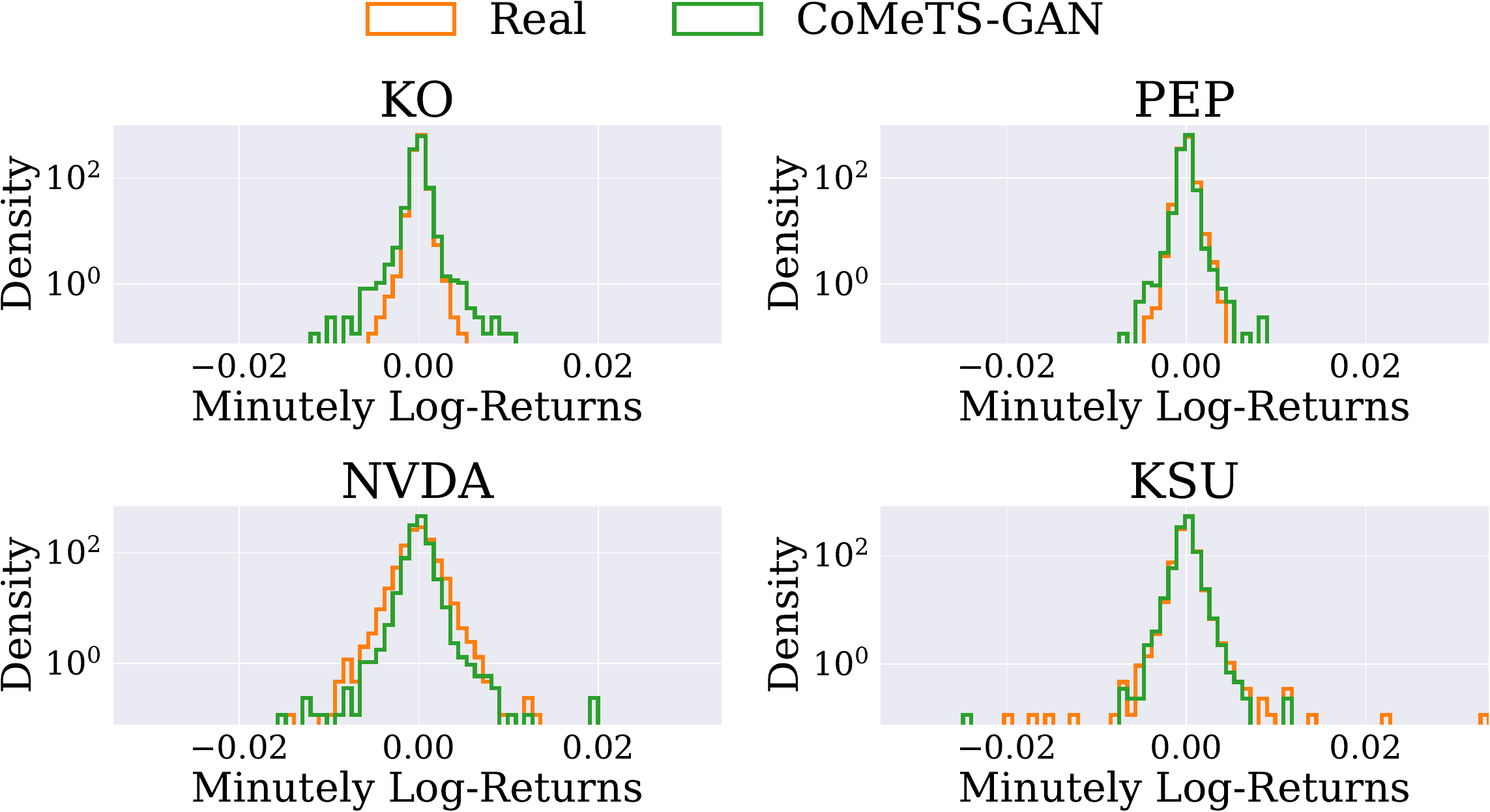}
        \caption{Intraday log-return distribution: $\Delta t = 1$.}
        \label{fig:sf_return_distrib}
    \end{subfigure}
    \begin{subfigure}{.49\linewidth}
        \centering
        \includegraphics[width=\linewidth]{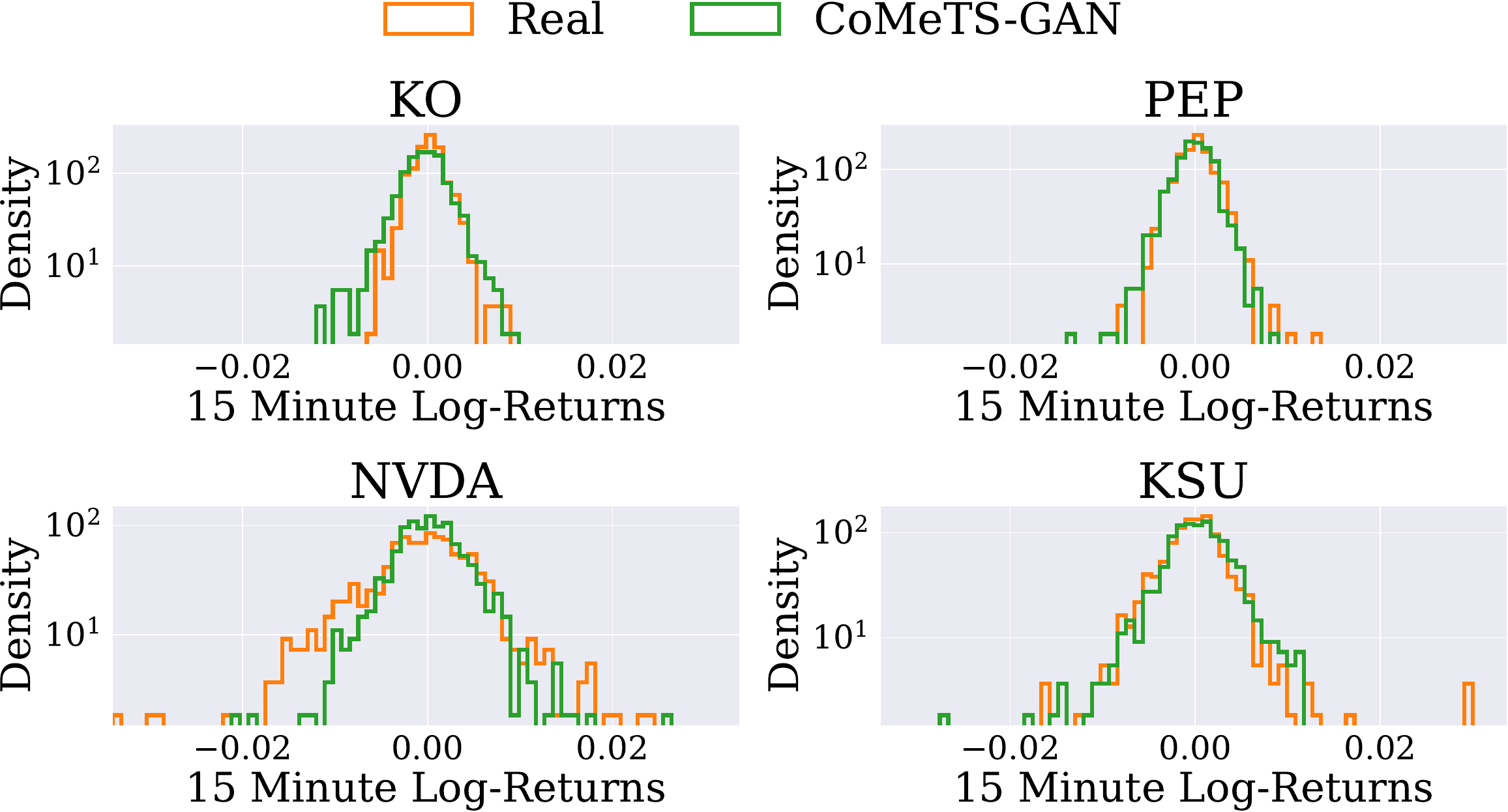}
        \caption{Intraday log-return distribution: $\Delta t = 15$.}
        \label{fig:sf_agg_gauss}
    \end{subfigure}
    \caption{Real and synthetic distribution of intraday log-return distributions.
    The similarity between the distributions assesses the model’s ability to replicate these statistical features of real market data.}
%\vspace{-.5cm}
\end{figure}

Figure~\ref{fig:sf_return_distrib} and ~\ref{fig:sf_agg_gauss} show the histogram of the intraday log-returns with periods of $\Delta t = 1$ and $\Delta t = 15$ minutes, respectively.
Results show a fat-tailed distribution as expected for the first one, and as we increase the period $\Delta t$, the distribution of returns better fits a normal distribution.
The described property is often referred to as \textit{aggregational normality}.
As seen in the figures, both real and generated data closely follow these known properties for all the stocks.
%\clearpage

\begin{figure}[h]
    \centering
    \begin{subfigure}{0.49\textwidth}
        \centering
        \includegraphics[width=\textwidth]{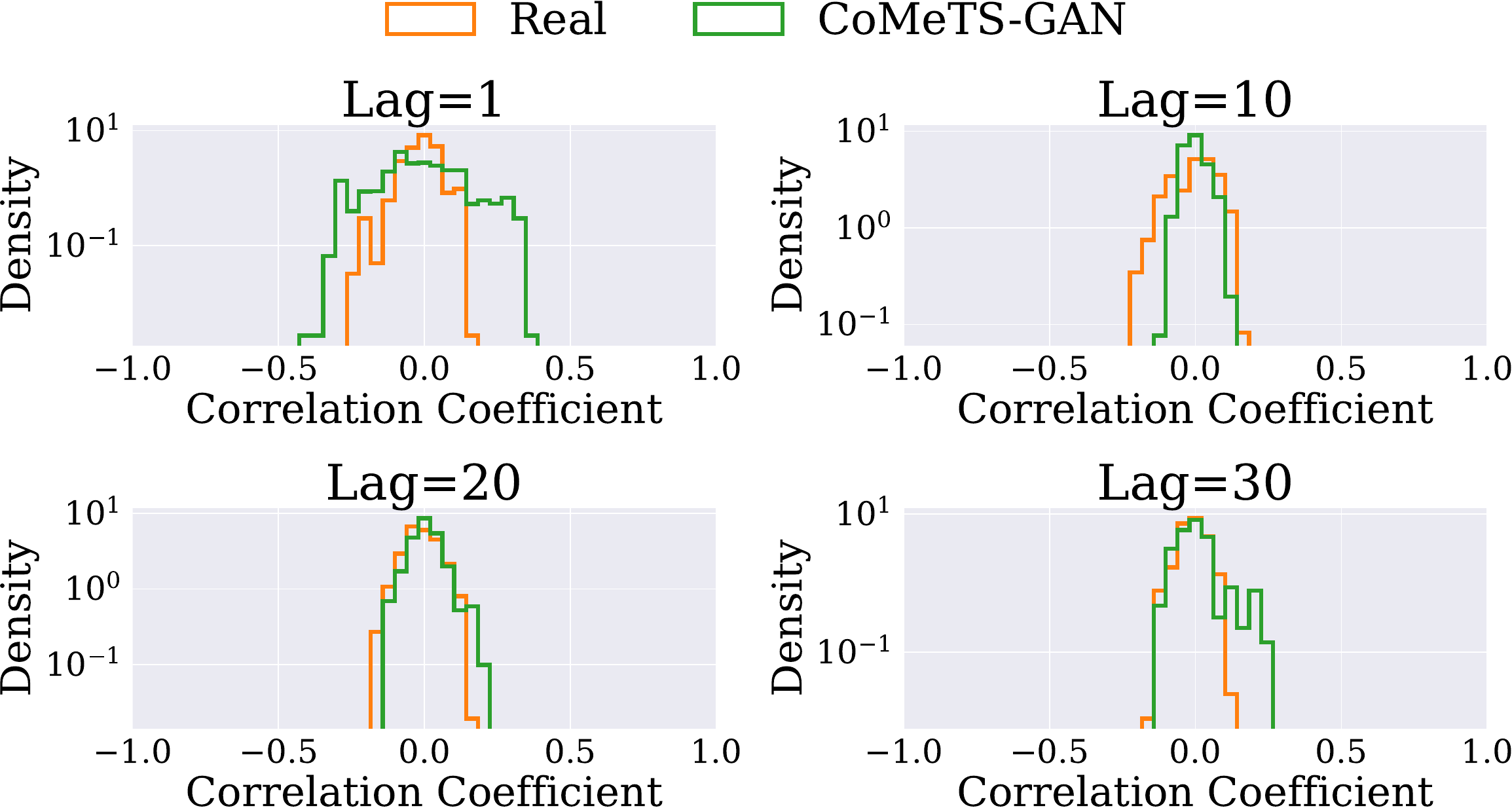}
        \caption{KO}    
        \label{fig:mean and std of net14}
    \end{subfigure}
    \hfill
    \begin{subfigure}{0.49\textwidth}  
        \centering 
        \includegraphics[width=\textwidth]{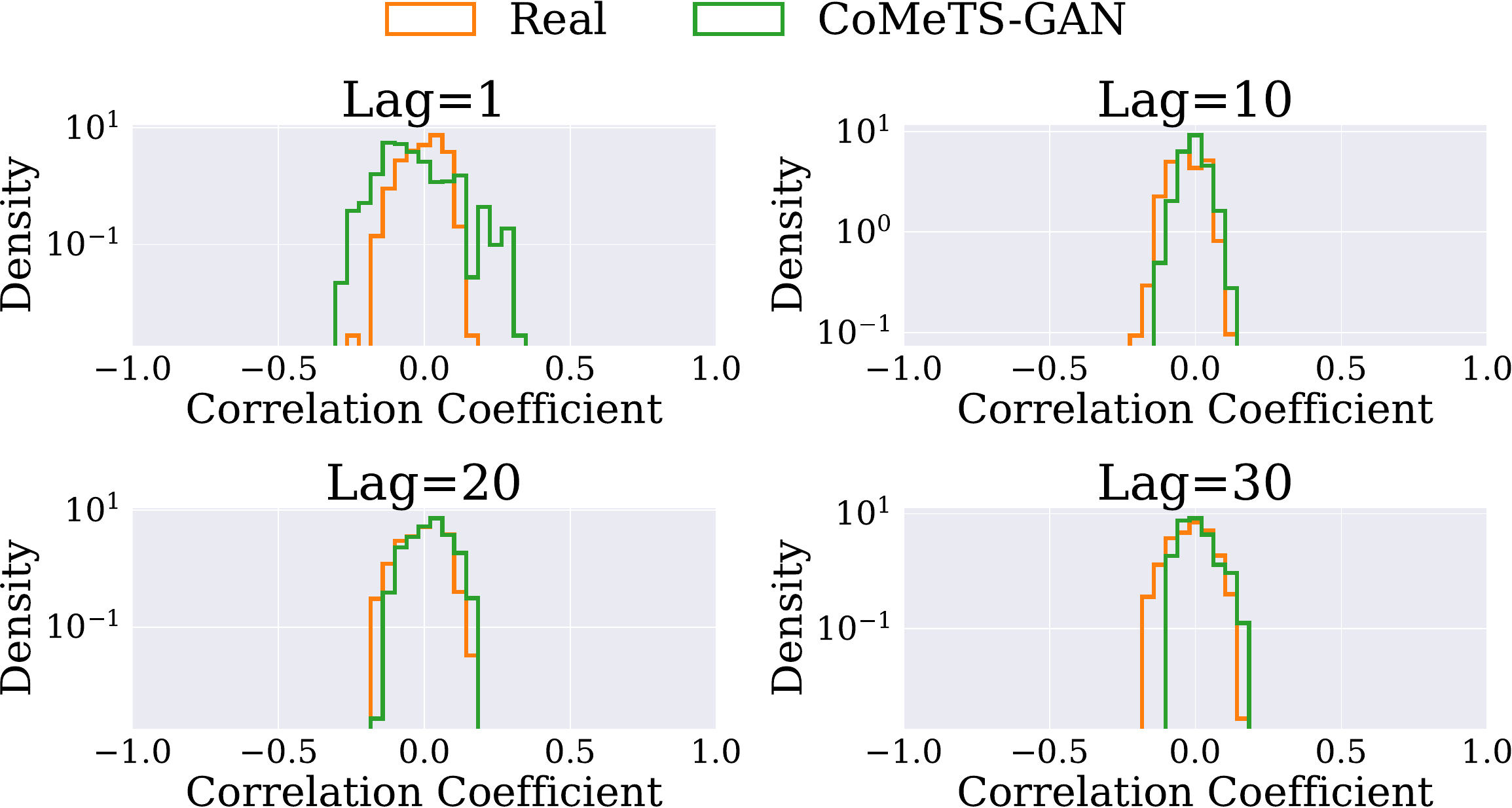}
        \caption{PEP}    
    \end{subfigure}
    %\vskip\baselineskip
    \begin{subfigure}{0.49\textwidth}   
        \centering 
        \includegraphics[width=\textwidth]{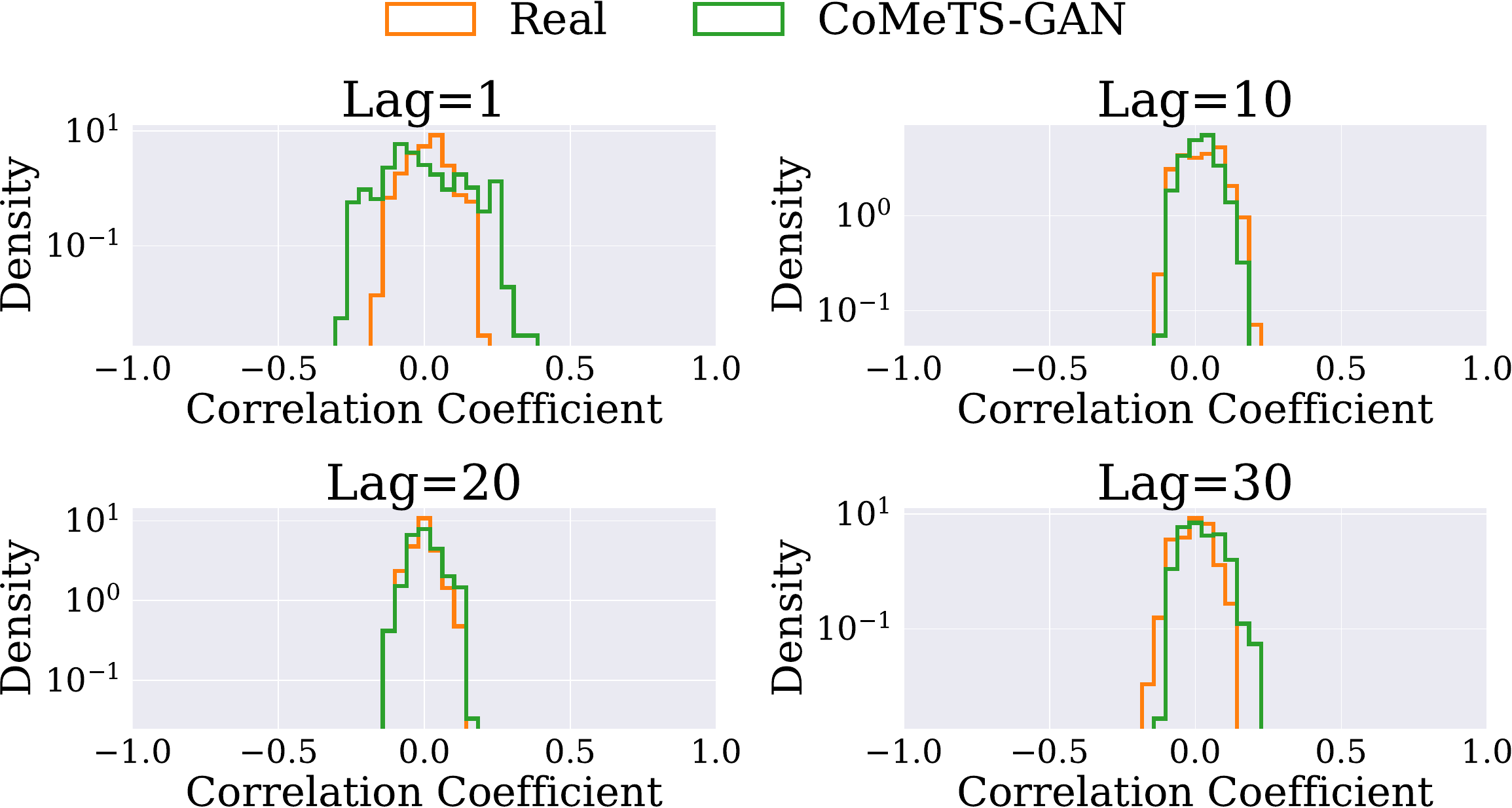}
        \caption{NVDA}    
    \end{subfigure}
    \hfill
    \begin{subfigure}{0.49\textwidth}   
        \centering 
        \includegraphics[width=\textwidth]{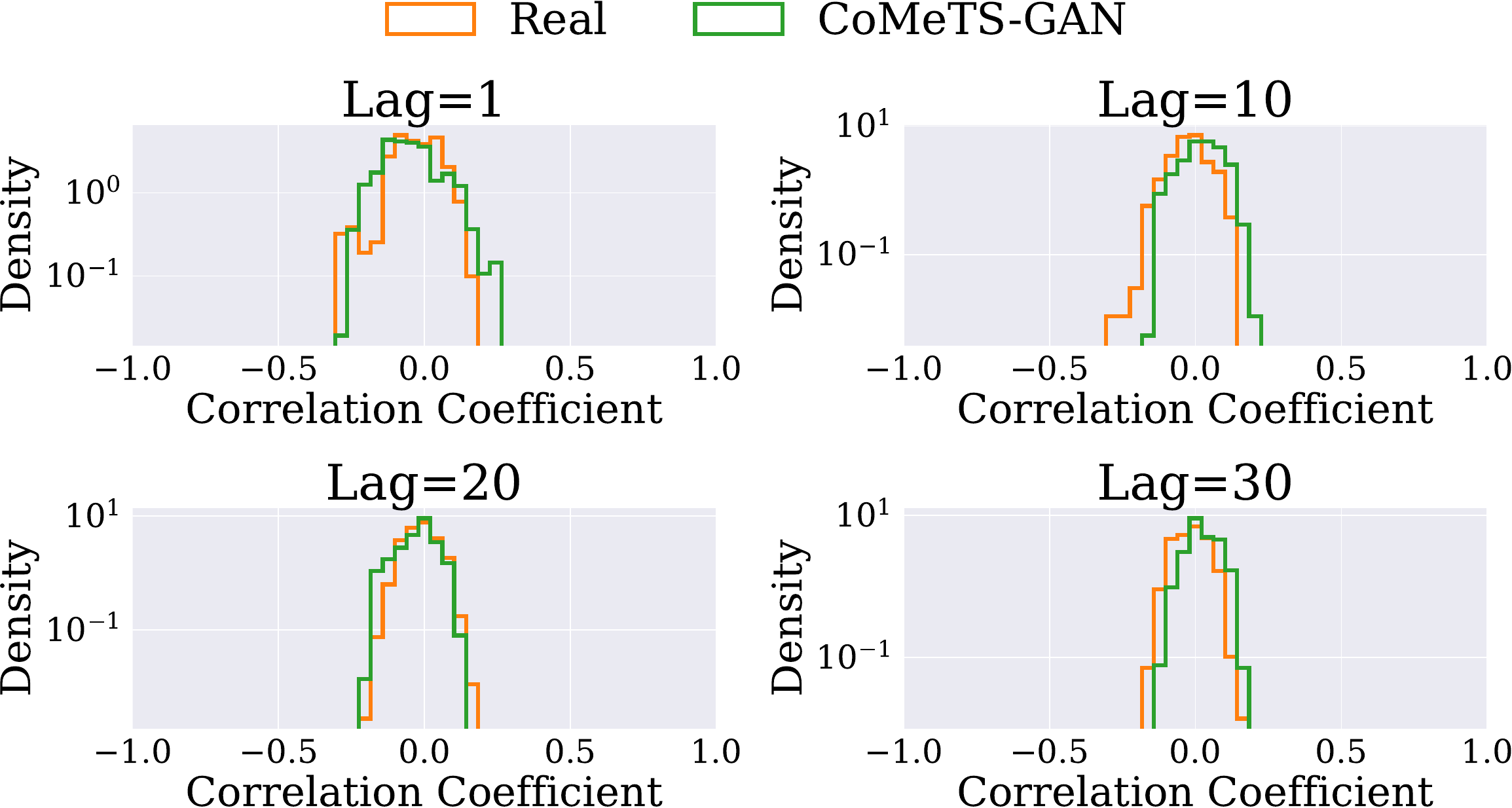}
        \caption{KSU}    
    \end{subfigure}
    \caption{%Autocorrelation of returns at increasing lags $\Delta t$ of 1, 10, 20, 30 minutes.
    Distributions of returns autocorrelation coefficients with increasing lags (\(\Delta t = 1, 10, 20, 30\) min). 
    The close match between real and CoMeTS-GAN demonstrates that the model faithfully replicates the diminishing autocorrelation.
    } 
    \label{fig:sf_autocorr}
\end{figure}

In Figure~\ref{fig:sf_autocorr}, we plot the autocorrelation of the minutely intraday log-returns of the stocks with different time lags of 1, 10, 20, and 30 minutes.
As we can observe, at increasing lag, the returns tend to cluster around a null correlation value.
The stylized fact related to the \textit{absence of autocorrelation} is thus present in both real and generated traces.

\begin{figure}[h]
    \centering
    \includegraphics[width=.7\linewidth]{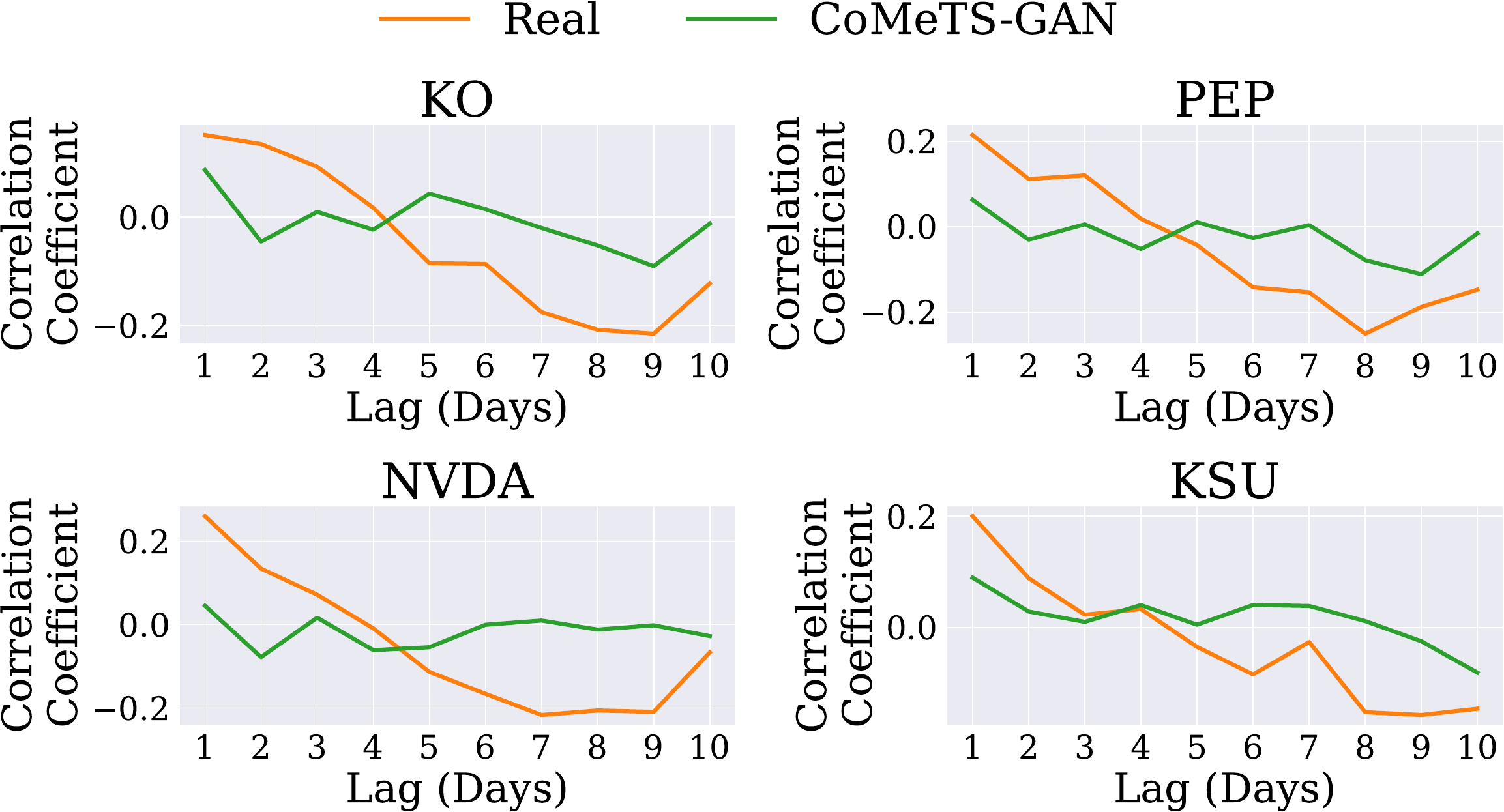}
    \caption{%Correlation of the volatility at increasing day lag.
    Correlation coefficients of volatility at increasing day lag. \framName successfully captures the general trend and low level of volatility autocorrelations found in real financial data.
    }
    \label{fig:sf_volatility_clustering}
    %\vspace{-1cm}
\end{figure}

Figure~\ref{fig:sf_volatility_clustering} shows the autocorrelation coefficients of the volatility at varying day lag.
The volatility is computed as the standard deviation of the returns.
We observe that the correlation is much higher when the lag is short and decreases as the lag increases, confirming that the volatility tends to cluster in time.
The plot also confirms that all the considered stocks show this behavior, both in the real and generated traces.
%\vspace{-.3cm}

\subsubsection{Comparison with State-of-the-Art}
We now compare \framName with several state-of-the-art approaches, namely TTS-GAN~\cite{li2022tts}, COSCI-GAN~\cite{seyfi2022generating}, GT-GAN~\cite{jeon2022gt}, TimeGAN~\cite{yoon2019time}, RCGAN~\cite{estebanRealvaluedMedicalTime2017} and C-RNN-GAN~\cite{mogrenCRNNGANContinuousRecurrent2016}.
For purely auto-regressive approaches, we compare against RNNs trained with teacher-forcing (T-Forcing)~\cite{graves2013generating} as well as professor-forcing (P-Forcing)~\cite{lamb2016professor}.
For additional comparison, we consider the performance of WaveNet~\cite{oord2016wavenet} as well as its GAN counterpart WaveGAN~\cite{donahue2018adversarial}.
We first focus on standard general-domain benchmarks from the literature, and the we focus on financial data.

\paragraph{Synthetic data realism on literature benchmarks}
We first show the results obtained on the Gaussian model dataset and on the Sines dataset in Table~\ref{tab:ds_gaussia} in terms of discriminative score (i.e., realism). 

\begin{table*}[b]
\renewcommand{\arraystretch}{.85}
\resizebox{1\textwidth}{!}{
\centering
\scriptsize
\begin{tabular}{@{}l|llllll|l@{}}
\toprule
\multicolumn{1}{c|}{} &\multicolumn{6}{c|}{Gaussian Dataset} &\multicolumn{1}{c}{Sines Dataset} \\
\cmidrule(l){2-8} 
\multicolumn{1}{c|}{} & \multicolumn{3}{c|}{Temporal Correlations (fixing $\sigma = 0.8$)} & \multicolumn{3}{c|}{Feature Correlations (fixing $\phi = 0.8$)} & \multicolumn{1}{c}{}\\ 
\cmidrule(l){2-8} 
\multicolumn{1}{c|}{Model}& \multicolumn{1}{c|}{$\phi = 0.2$}& \multicolumn{1}{c|}{$\phi = 0.5$}& \multicolumn{1}{c|}{$\phi = 0.8$}& \multicolumn{1}{c|}{$\sigma = 0.2$}& \multicolumn{1}{c|}{$\sigma = 0.5$}& \multicolumn{1}{c|}{$\sigma = 0.8$} & \multicolumn{1}{c}{} \\ 
\cmidrule(l){2-8} 
\multicolumn{1}{c|}{} & \multicolumn{7}{c}{Discriminative score (lower the better)} \\ 
\midrule
\framName (w/o cross-corr.) & \multicolumn{1}{l|}{$0.211 \pm 0.007$}          & \multicolumn{1}{l|}{$0.195 \pm 0.005$}          & \multicolumn{1}{l|}{$0.137 \pm 0.012$}          & \multicolumn{1}{l|}{$0.187 \pm 0.006$}          & \multicolumn{1}{l|}{$0.206 \pm 0.018$}          & $0.137 \pm 0.012$                               & $--$ \\
\framName (w/ cross-corr.)  & \multicolumn{1}{l|}{$0.187 \pm 0.003$}          & \multicolumn{1}{l|}{$0.183 \pm 0.005$}          & \multicolumn{1}{l|}{$\underline{0.102} \pm 0.006$}          & \multicolumn{1}{l|}{$0.180 \pm 0.009$}          & \multicolumn{1}{l|}{$0.164 \pm 0.011$}          & $\underline{0.102} \pm 0.006$                               & $0.013 \pm 0.007$ \\
TTS-GAN                     & \multicolumn{1}{l|}{$0.494 \pm 0.004$}          & \multicolumn{1}{l|}{$0.470 \pm 0.010$}          & \multicolumn{1}{l|}{$0.437 \pm 0.025$}          & \multicolumn{1}{l|}{$0.305 \pm 0.019$}          & \multicolumn{1}{l|}{$0.311 \pm 0.037$}          & \multicolumn{1}{l|}{$0.437 \pm 0.024$}          & $0.092 \pm 0.045$ \\
COSCI-GAN                   & \multicolumn{1}{l|}{$\mathbf{0.167} \pm 0.007$} & \multicolumn{1}{l|}{$\mathbf{0.105} \pm 0.013$} & \multicolumn{1}{l|}{$\mathbf{0.070} \pm 0.018$} & \multicolumn{1}{l|}{$\mathbf{0.091} \pm 0.020$} & \multicolumn{1}{l|}{$\mathbf{0.075} \pm 0.030$} & \multicolumn{1}{l|}{$\mathbf{0.070} \pm 0.018$} & $0.042 \pm 0.019$ \\
GT-GAN                      & \multicolumn{1}{l|}{$0.364 \pm 0.082$}          & \multicolumn{1}{l|}{$\underline{0.162} \pm 0.122$}          & \multicolumn{1}{l|}{$0.172 \pm 0.120$}          & \multicolumn{1}{l|}{$\underline{0.085} \pm 0.091$}          & \multicolumn{1}{l|}{$\underline{0.147} \pm 0.110$}          & \multicolumn{1}{l|}{$0.172 \pm 0.120$}          & $\underline{0.012} \pm 0.014$ \\
TimeGAN                     & \multicolumn{1}{l|}{$\underline{0.175} \pm 0.006$}          & \multicolumn{1}{l|}{$0.174 \pm 0.012$}          & \multicolumn{1}{l|}{$0.105 \pm 0.005$}          & \multicolumn{1}{l|}{$0.181 \pm 0.006$}          & \multicolumn{1}{l|}{$0.152 \pm 0.011$}          & $0.105 \pm 0.005$                               & $\mathbf{0.011} \pm 0.008$ \\
RCGAN                       & \multicolumn{1}{l|}{$0.177 \pm 0.012$}          & \multicolumn{1}{l|}{$0.190 \pm 0.011$}          & \multicolumn{1}{l|}{$0.133 \pm 0.019$}          & \multicolumn{1}{l|}{$0.186 \pm 0.012$}          & \multicolumn{1}{l|}{$0.190 \pm 0.012$}          & $0.133 \pm 0.019$                               & $0.022 \pm 0.008$ \\
C-RNN-GAN                   & \multicolumn{1}{l|}{$0.391 \pm 0.006$}          & \multicolumn{1}{l|}{$0.227 \pm 0.017$}          & \multicolumn{1}{l|}{$0.220 \pm 0.016$}          & \multicolumn{1}{l|}{$0.198 \pm 0.011$}          & \multicolumn{1}{l|}{$0.202 \pm 0.010$}          & $0.220 \pm 0.016$                               & $0.229 \pm 0.040$\\
T-Forcing                   & \multicolumn{1}{l|}{$0.500 \pm 0.000$}          & \multicolumn{1}{l|}{$0.500 \pm 0.000$}          & \multicolumn{1}{l|}{$0.499 \pm 0.001$}          & \multicolumn{1}{l|}{$0.499 \pm 0.001$}          & \multicolumn{1}{l|}{$0.499 \pm 0.001$}          & $0.499 \pm 0.001$                               & $0.495 \pm 0.001$\\
P-Forcing                   & \multicolumn{1}{l|}{$0.498 \pm 0.002$}          & \multicolumn{1}{l|}{$0.472 \pm 0.008$}          & \multicolumn{1}{l|}{$0.396 \pm 0.018$}          & \multicolumn{1}{l|}{$0.460 \pm 0.003$}          & \multicolumn{1}{l|}{$0.408 \pm 0.016$}          & $0.396 \pm 0.018$                               & $0.430 \pm 0.027$ \\
WaveNet                     & \multicolumn{1}{l|}{$0.337 \pm 0.005$}          & \multicolumn{1}{l|}{$0.235 \pm 0.009$}          & \multicolumn{1}{l|}{$0.229 \pm 0.013$}          & \multicolumn{1}{l|}{$0.217 \pm 0.010$}          & \multicolumn{1}{l|}{$0.226 \pm 0.011$}          & $0.229 \pm 0.013$                               & $0.158 \pm 0.011$\\
WaveGAN                     & \multicolumn{1}{l|}{$0.336 \pm 0.011$}          & \multicolumn{1}{l|}{$0.213 \pm 0.013$}          & \multicolumn{1}{l|}{$0.230 \pm 0.023$}          & \multicolumn{1}{l|}{$0.192 \pm 0.012$}          & \multicolumn{1}{l|}{$0.205 \pm 0.015$}          & $0.230 \pm 0.023$                               & $0.277 \pm 0.013$\\ 
\bottomrule
\end{tabular}}

\caption{Discriminative score for the Gaussian and Sines datasets. First and second best are highlighted in bold and underline respectively.}
\label{tab:ds_gaussia}
\vspace{-.3cm}
\end{table*}

Numerical results show that our model performances are somehow comparable with respect to more complex state-of-the-art approaches. In fact, our model is specifically designed for financial data while the literature benchmarks do not focus on financial datasets. This experiment highlights both the potential performance and the versatility of our model, beyond the specific domain for which it was originally developed.

It is also important to note that our model is structurally different from more advanced models such as TimeGAN, COSCI-GAN, and GT-GAN. In such work, the original \ts is factorized in short sub-sequences, and the model outputs a synthetic set of such sub-sequences at inference time, but there is no natural and intuitive way to recombine them to obtain a longer sequence, retaining temporal continuation. Instead, our model is able to generate a single \ts with up to 9360 observations (24 trading days).
Furthermore, the training times of such complex models are generally higher --- typically in the range of tens of hours (e.g., TimeGAN required 39 hours of training, compared to just 4 hours and 20 minutes for our model).

Finally, Table~\ref{tab:ds_gaussia}
presents also an ablation study to assess the contribution of the cross-correlation term in the critic score. In particular, rows one and two of the table report the discriminative score for $\alpha = 0$ and $\alpha = 1$, respectively. The results indicate that incorporating the cross-correlation term improves model performance, especially in settings with high inter-feature correlation (i.e., high values of $\phi$), thereby demonstrating the effectiveness of the cross-correlation coefficients.

\paragraph{Correlation dynamics on financial synthetic data}
As previously mentioned, our model has comparable performance wrt more recent approaches, though it does not outperform them in terms of realism. However, our model is specifically designed for financial data, and to improved correlation dynamics of synthetic data. 
Therefore, we now evaluate the most promising approaches in Table~\ref{tab:ds_gaussia}, i.e. COSCI-GAN~\cite{seyfi2022generating} and GT-GAN~\cite{jeon2022gt}, with respect to our financial dataset, to highlight the advantages of \framName over the aforementioned general-purpose approaches.

\begin{figure}
    \centering
    \includegraphics[width=.7\linewidth]{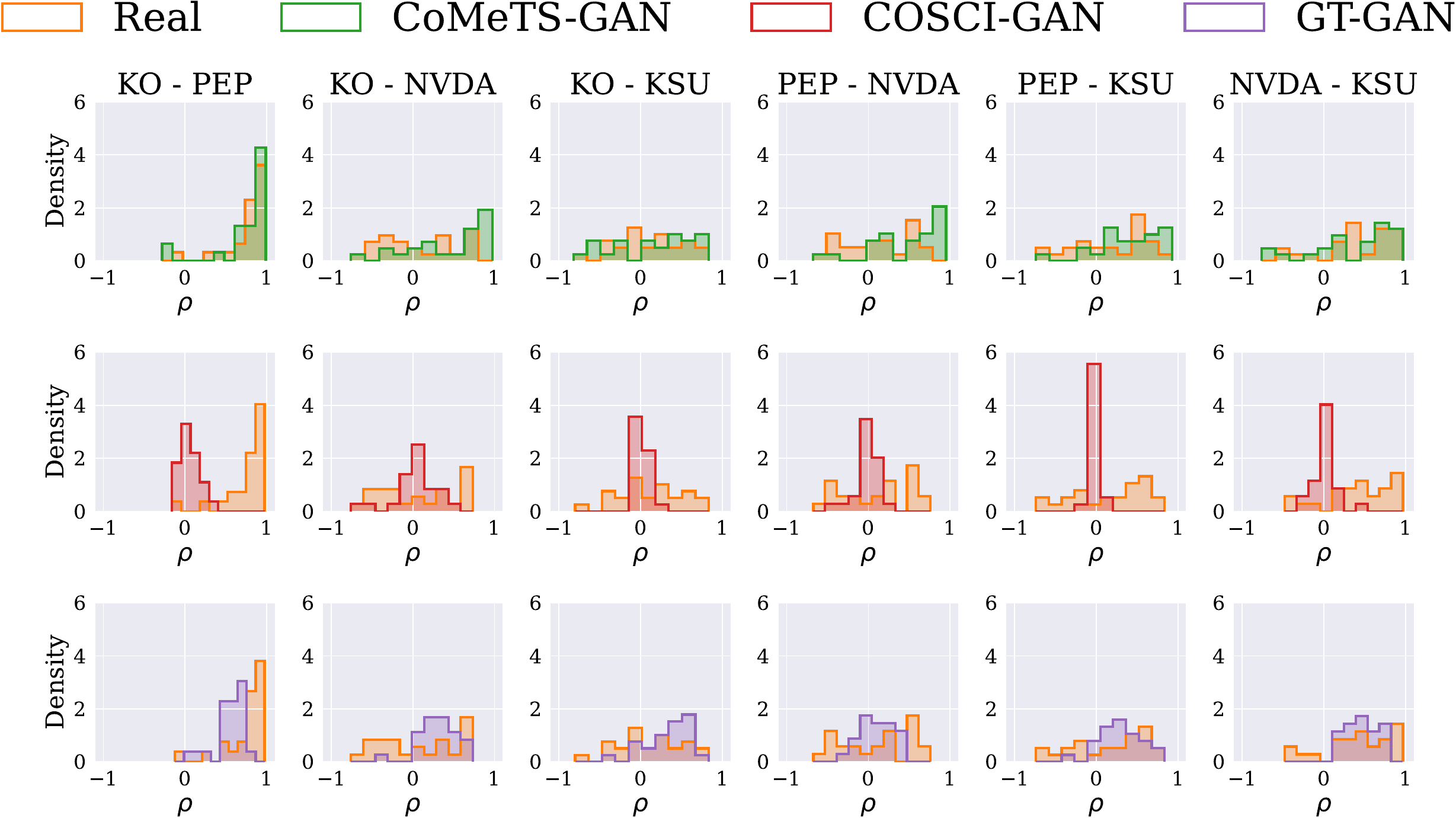}
    \caption{Pairwise correlation distributions of daily asset prices (390 minutes). 
    \framName most closely matches the empirical correlation structures.%, demonstrating its effectiveness in capturing complex cross-asset dependencies.
    }
    \label{fig:correlations_sota}
    %\vspace{-.5cm}
\end{figure}

Figure~\ref{fig:correlations_sota} depicts the pair-wise correlations of the stocks' price over windows of one trading day (390 time-steps) showing that \framName is the best model in respecting the correlation values of the real data, as also reported in Table~\ref{tab:correlations_sota_wass} in terms of Wasserstein distance between the histograms.
\begin{figure}[h]
\centering
\begin{tabular}{l|c|c|c}
\toprule
 & \framName & COSCI-GAN & GT-GAN \\
\midrule
KO - PEP & \textbf{0.13} & 0.29 & 0.30 \\
KO - NVDA & \textbf{0.19} & 0.28 & 0.28 \\
KO - KSU & \textbf{0.10} & 0.71 & 0.25 \\
PEP - NVDA & \textbf{0.20} & 0.52 & 0.29 \\
PEP - KSU & \textbf{0.25} & 0.85 & 0.26 \\
NVDA - KSU & \textbf{0.05} & 0.52 & 0.35 \\
\bottomrule
\end{tabular}
\captionof{table}{Wasserstein distance of the distributions in Figure~\ref{fig:correlations_sota}.
}
\label{tab:correlations_sota_wass}
\end{figure}

\begin{figure}[h]
    \centering
    \includegraphics[width=.7\linewidth]{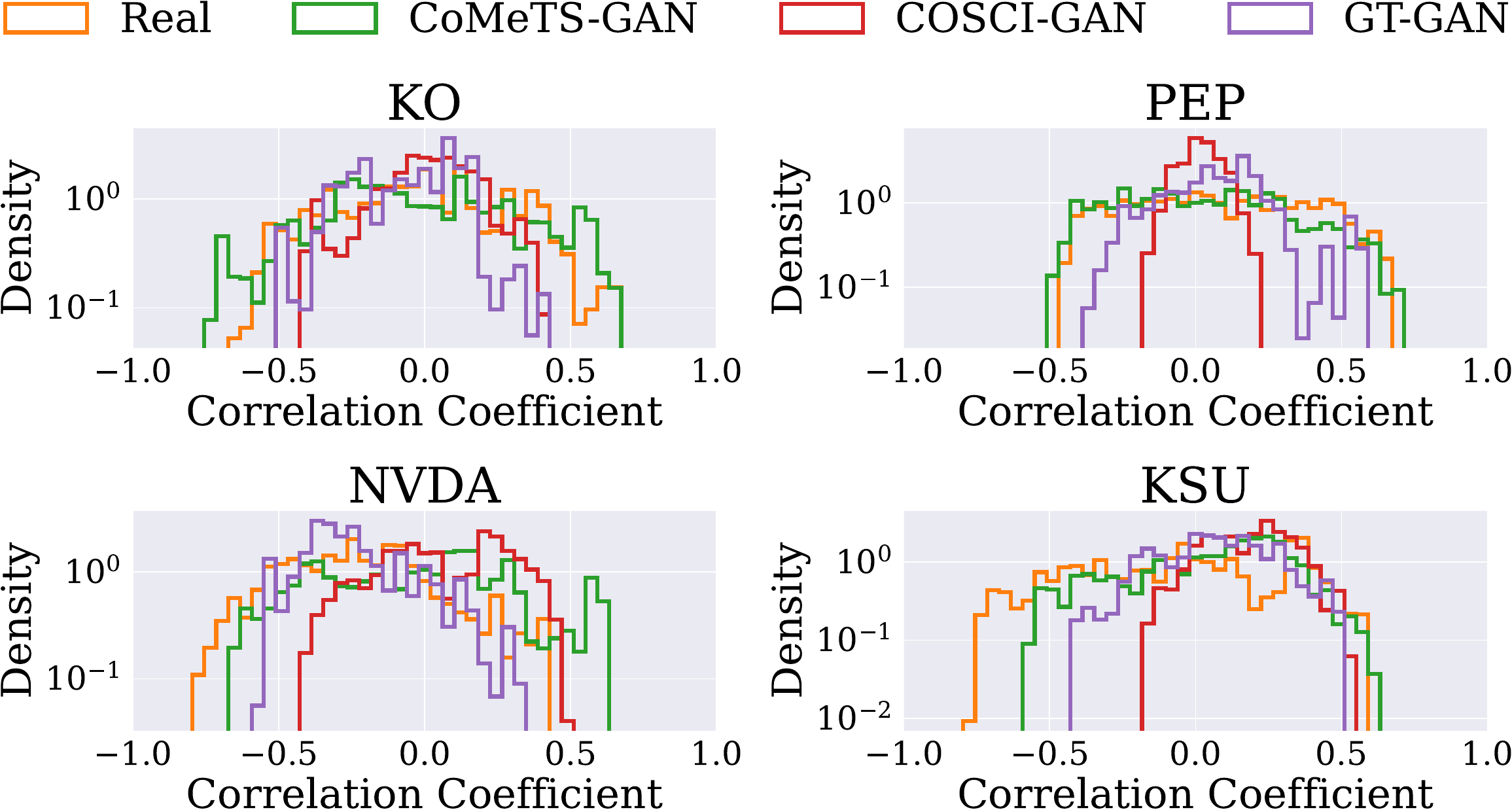}
    \caption{%Volume-volatility correlation.
    Distributions of volume–volatility correlation coefficients over windows of two days. \framName shows superior ability to reproduce this key market relationship.% compared to COSCI-GAN and GT-GAN.
    }
    \label{fig:sf_volume_volatility_corr}
    %\vspace{-1cm}
\end{figure}

Figure~\ref{fig:sf_volume_volatility_corr} shows the distribution of the correlation between the average traded volume and the volatility over two trading days.
According to the stylized fact, trading volume and volatility are positively correlated, and it is clear that \framName's data are the best in reproducing this expected behavior. 
%\vspace{-.2cm}

\subsubsection{Reactivity}
An important what-if question in finance concerns the market impact of meta-orders and the subsequent propagation of price shocks~\cite{bouchaud2018trades}. In other words, how would the market react if a given amount of money were traded? In this section, we demonstrate how our conditional framework can be leveraged to simulate interactive market dynamics, and address such questions. In particular, we conducted an experiment in which a perturbation was manually introduced to the \ts of KO, to observe how the model reacts in generating the other stocks.
In particular, we focus on the reaction of PEP, which is highly positively correlated with KO. 

Specifically, let $\hat{\bm{x}}_{t:t+F}$ be the sub-sequence generated by the model starting at time $t$.
A perturbation is introduced by substituting what the model actually generates with $\hat{\bm{x}}_{t:t+F} + \alpha \cdot \sigma(\hat{\bm{x}}_{t:t+F})$, where $\sigma(\cdot)$ being the standard deviation function.

\begin{figure}[h]
    \centering
    \includegraphics[width=.7\linewidth]{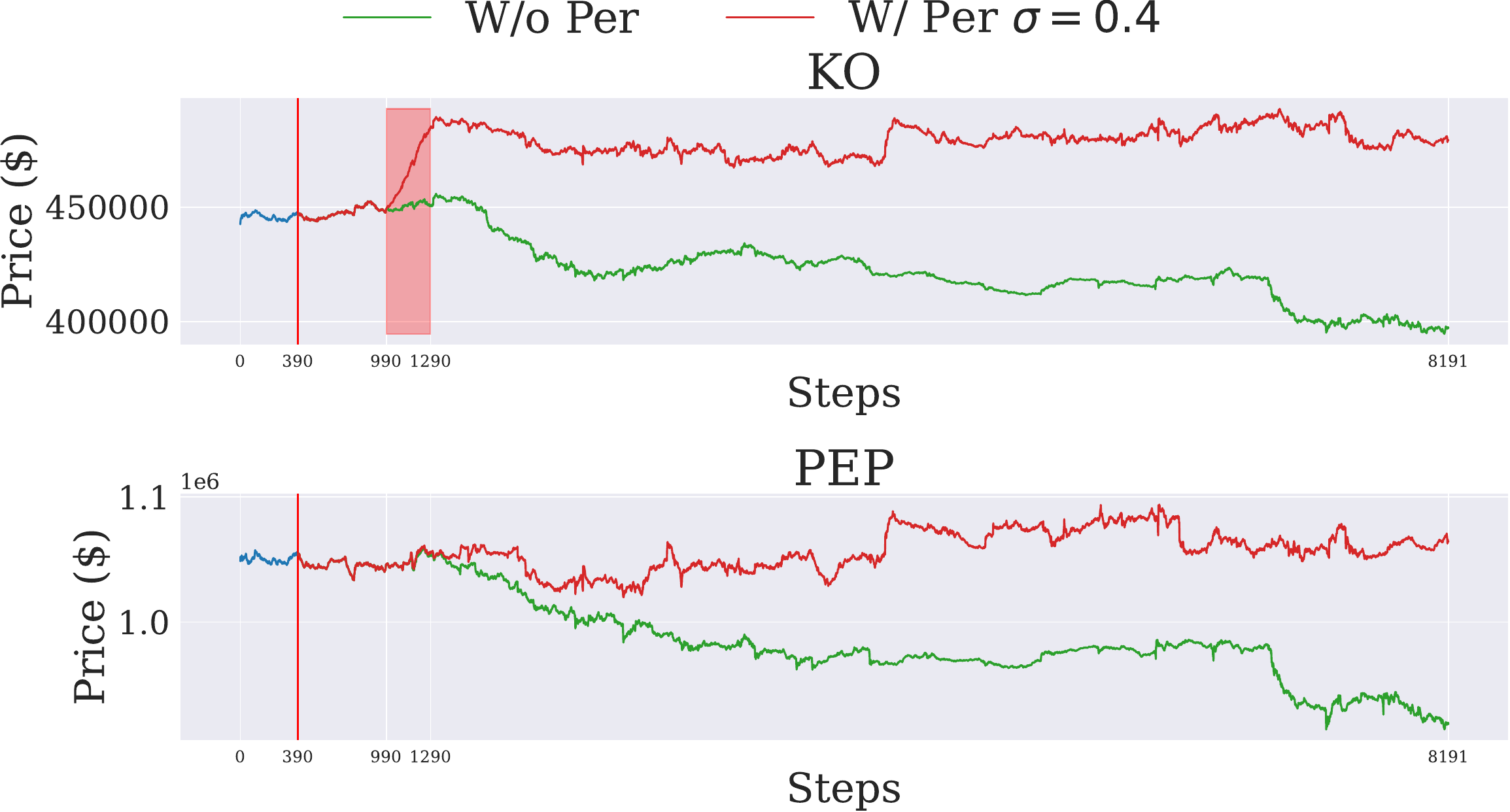}
    \caption{%Perturbation of KO and the reaction of PEP.
    Price evolution of KO (top) and PEP (bottom) under two scenarios: with and without a perturbation applied to KO ($\sigma = 0.4$, highlighted period). 
    }
    \label{fig:introducing_perturbations}
    %\vspace{-.8cm}
\end{figure}

Figure~\ref{fig:introducing_perturbations} shows the perturbation introduced on KO in the red window.
It is interesting to observe how the model ``propagates" the shock and adjusts the generation of the non-perturbed stock in order to preserve the correlation following the perturbation event. 
%\vspace{-.5cm}

\begin{figure}[h]
    \centering
    \includegraphics[width=.7\linewidth]{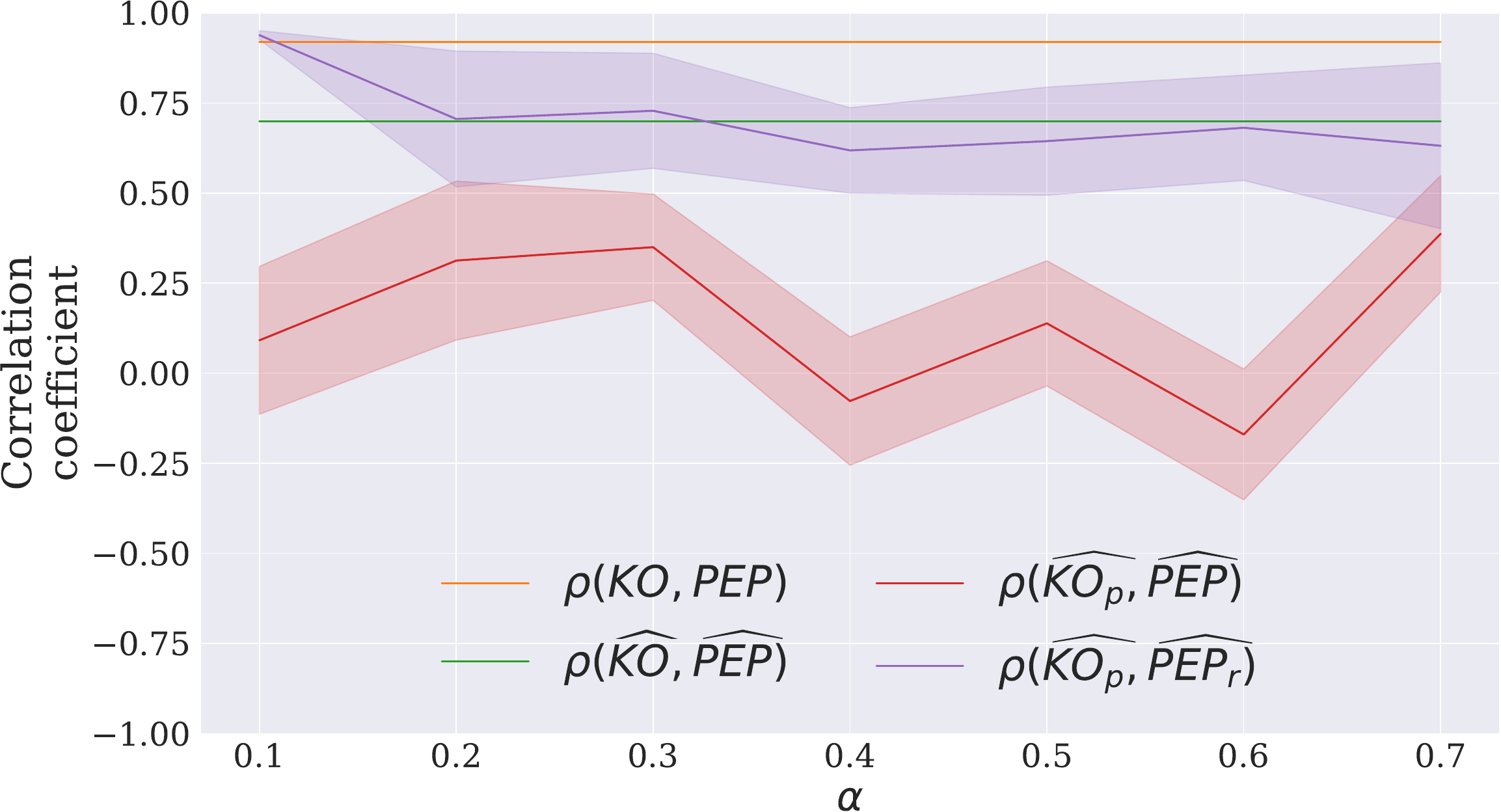}
    \caption{Correlation of the KO and PEP stocks varying the intensity of the perturbation.
    The experiment demonstrates that a shock to KO’s price induces a correlated response in PEP, illustrating \framName’s capacity to simulate realistic cross-asset market reactions, preserving inter-stock dependencies following external interventions.
    }
    \label{fig:reactive_correlations}
    %\vspace{-.5cm}
\end{figure}

Figure~\ref{fig:reactive_correlations} shows the correlation of KO and PEP, varying the intensity of the perturbation.
The orange line is the correlation between real KO and real PEP, which is highly positive, close to 1.
The green line is the correlation between their synthetic versions $\widehat{\text{KO}}$ and $\widehat{\text{PEP}}$, which is still positive and high. 
Let $\widehat{\text{KO}_p}$ be the synthetic version of KO subject to the perturbation.
The red line is the correlation between $\widehat{\text{KO}_p}$ and $\widehat{\text{PEP}}$.
Let $\widehat{\text{PEP}_r}$ be the synthetic version of PEP generated by the model in response to the perturbation observed in KO.
The purple line is the correlation between $\widehat{\text{KO}_p}$ and $\widehat{\text{PEP}_r}$.
The experiment was repeated for several seeds, generating the standard error evidenced by the bands in the figure.
As we can see, the model punctually intervenes in adjusting the generation of PEP to preserve the correlation with KO after the perturbation.

\subsubsection{Scalability.}
\begin{figure}
    \centering
    \includegraphics[width=\linewidth]{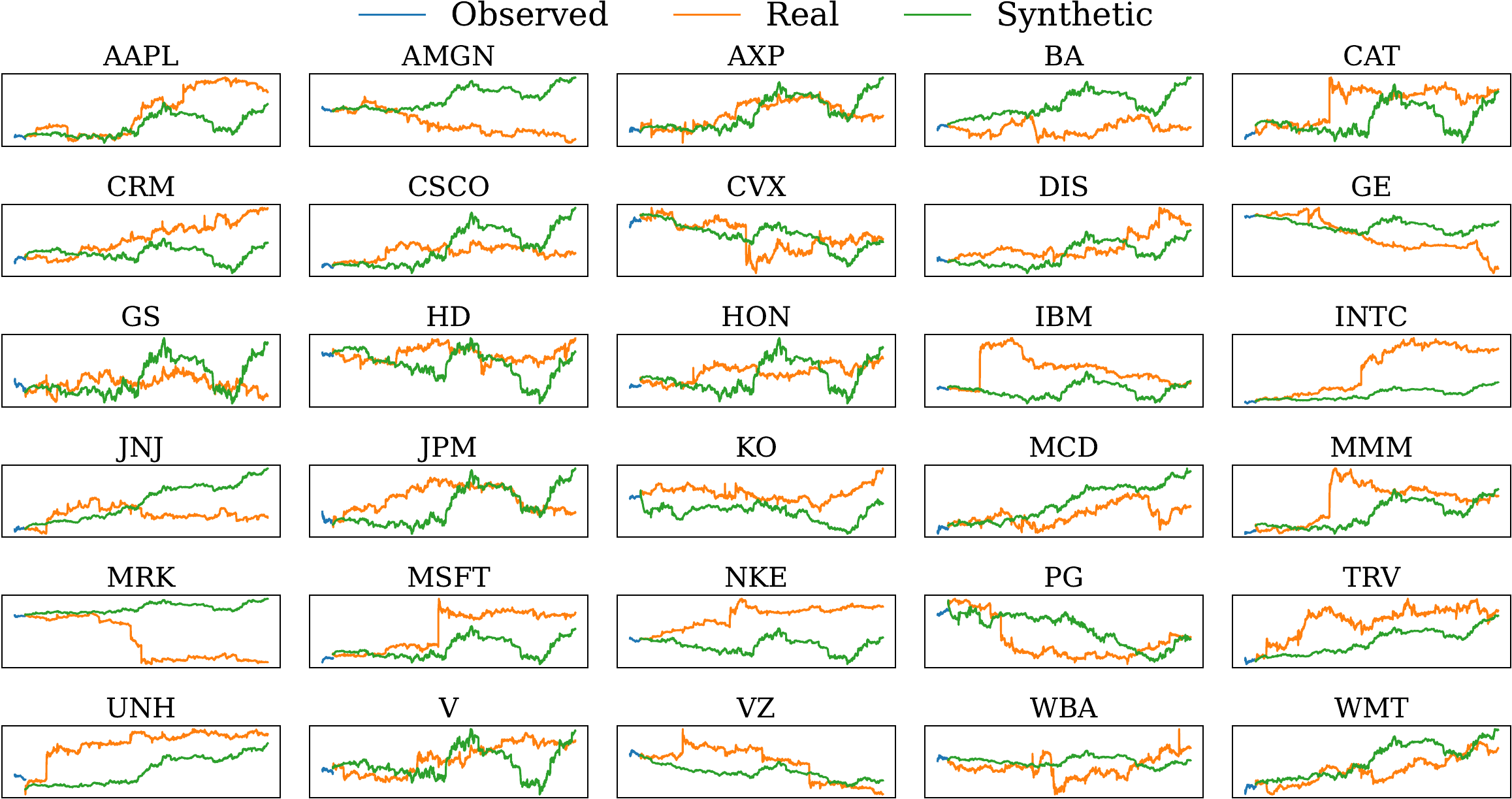}
    \caption{Concurrent price generation of the 30 components of the DJIA, demonstrating the framework’s scalability.}
    \label{fig:DJIA_prices}
\end{figure}

Considering the model's architecture in Figure~\ref{fig:gan_architecture}, the size of the input and output of the generator and the input of the discriminator linearly depend on $n$.
On the other hand, the intermediate hidden layers of the networks remain constant with respect to $n$.
The linear layer in the critic takes as input a vector of size $\binom{2n}{2}$. 
To test the model's scalability, we run the auto-regressive application on all 30 stocks included in the DJIA index (Figure~\ref{fig:DJIA_prices}), obtaining realistic correlation dynamics. 
Indeed, the synthetic traces still show the expected correlation properties, even when considered in the simultaneous generation of numerous \ts.
Furthermore, all the stylized facts that characterize the returns are preserved.

\subsection{A Critic-as-a-Guide framework}
While WGANs leverage a critic to estimate the Wasserstein distance and enforce realistic data distributions, diffusion models utilize iterative noise conditioning to generate high-fidelity samples. 
In this section, we explore the theoretical and practical implications of repurposing a pre-trained WGAN critic as a guidance mechanism for classifier-guided diffusion models, with a focus on stock price generation. 

\begin{figure}
    \centering
    \begin{subfigure}{\linewidth}
        \centering
        \includegraphics[width=0.8\linewidth]{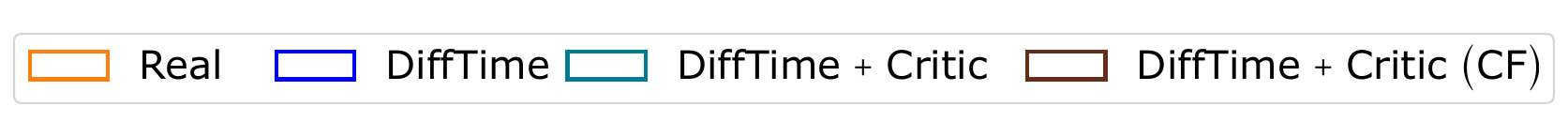}
    \end{subfigure}
    \begin{subfigure}{\linewidth}
        \centering        \includegraphics[width=\linewidth]{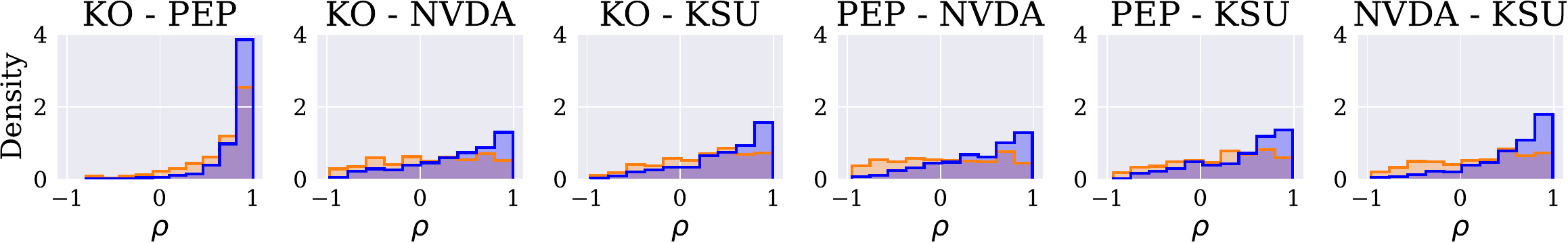}
        %\vspace{-.6cm}
        \caption{Synthetic.}
        \label{fig:correlations_synthetic}
    \end{subfigure}
    \begin{subfigure}{\linewidth}
        \centering

        \includegraphics[width=\linewidth]{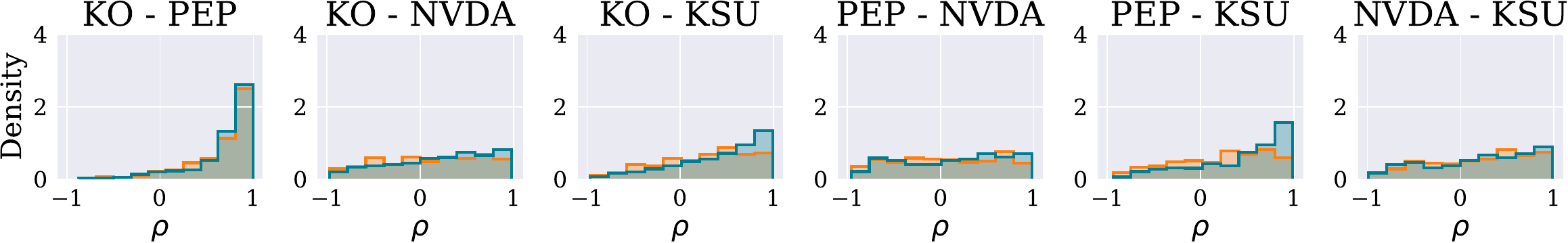}
        %\vspace{-.6cm}
        \caption{Synthetic with guidance.}
        \label{fig:correlations_guidance}
    \end{subfigure}
    \begin{subfigure}{\linewidth}
        \centering

        \includegraphics[width=\linewidth]{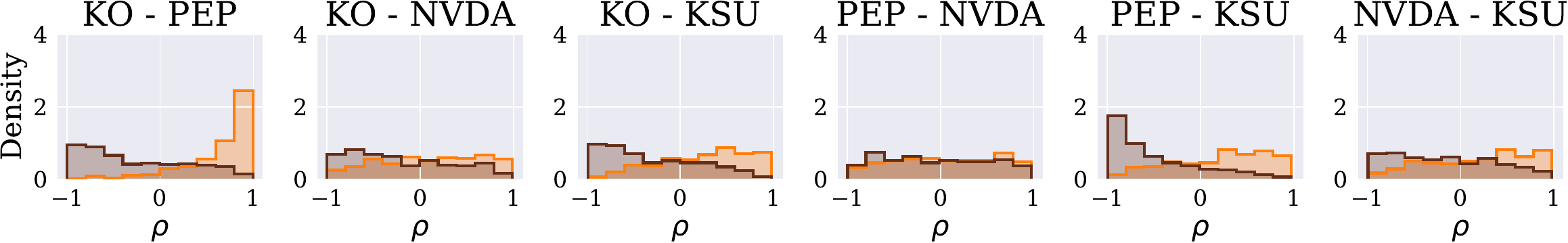}
        %\vspace{-.6cm}
        \caption{Synthetic with guidance for counterfactual generation.}
        \label{fig:correlations_counterfactual}
    \end{subfigure}
    %\vspace{-.6cm}
    \caption{%Prices correlations  - Diffusion models.
    Distributions of price correlations between asset pairs for real data and synthetic series generated by three diffusion model variants: \textit{DiffTime}, \textit{DiffTime + Critic}, and \textit{DiffTime + Critic (CF)}. 
    Adding the Critic as guidance (middle row) results in synthetic correlations that more closely match the real market structure, while counterfactual guidance (bottom row) produces distinctly different correlation patterns, illustrating the model’s flexibility for simulating both realistic and counterfactual scenarios.
    }
    %\vspace{-.4cm}
    \label{fig:correlations}
\end{figure}

To assess efficacy, we propose benchmarking both guided and unguided diffusion models by examining the correlation dynamics of the generated \ts.
Specifically, we employ a diffusion model based on DiffTime~\cite{coletta2023constrained} for the unconditional generation windows of 150 time-steps across four selected stocks: KO, PEP, NVDA, and KSU. 
For each generated window, we compute the pairwise correlations between the stocks, resulting in the distributions visualized in Figure~\ref{fig:correlations_synthetic}.
It illustrates the correlation distributions derived from the unguided diffusion model ($w=0$ in Eq.~\ref{eq:guidance}), while Figure~\ref{fig:correlations_guidance} shows those obtained with guided generation ($w=250$), highlighting a much closer alignment with the real data distribution and demonstrating a substantial improvement due to guidance.
To quantify the improvement, we report the Wasserstein distance in Table~\ref{tab:wasserstein_distance}.

To further highlight the value of re-purposing the WGAN critic to guide the generation, Figure~\ref{fig:correlations_counterfactual} presents the generation of a counter-factual scenario. In particular, the guidance is applied in the opposite direction, i.e., by flipping the classifier gradients ($w=-250$). 
By reversing the guidance signal, the generated samples are explicitly discouraged from aligning with the real data distribution, resulting in correlation patterns that diverge markedly from the empirical ones and from those generated by both the guided and unguided models. 
This experiment underscores the importance of effective guidance not only for enhancing sample quality but also for generating meaningful counterfactuals under adversarial guidance.

\begin{figure}[h]
\centering
\begin{tabular}{l|c|c}
\toprule
 & DiffTime + Critic ($w=250$) & DiffTime ($w=0$) \\
\midrule
KO - PEP & 0.07 & 0.26 \\
KO - NVDA & 0.09 & 0.21 \\
KO - KSU & 0.15 & 0.20 \\
PEP - NVDA & 0.04 & 0.23 \\
PEP - KSU & 0.17 & 0.19 \\
NVDA - KSU & 0.04 & 0.30 \\
\bottomrule
\end{tabular}
\captionof{table}{Wasserstein distance of the distributions in Figure~\ref{fig:correlations}.}
\label{tab:wasserstein_distance}
%\vspace{-.4cm}
\end{figure}

\section{Discussion and Conclusion}
\label{sec:conclusion}
In this work, we discuss novel solutions for generating high-quality interdependent markets using deep learning models. 
Our quality-aware framework introduces three key contributions: First, we create a new critic score for a C-WGAN, specifically suitable for \ts generation, based on the degree of realism of the cross-correlation between the features; Then we introduce a new architecture based on a C-WGAN, to capture aspects of interdependence among financial markets and to generate \ts with similar characteristics to real ones; Finally, we discuss how our Critic can be repurposed to enhance the performance of any existing diffusion model framework.
We thoroughly analyze the model's performance both on benchmark datasets and new stock market traces measuring the performance in terms of \textit{Discriminative Score}, \textit{Diversity}, \textit{Stylized Facts}, \textit{Reactivity}, and \textit{Scalability}.
The results are promising: in all respects, the proposed architecture yields good results with respect to state-of-the-art approaches such as TimeGAN, solving its inherent issues related to training time and temporal continuation of long \ts. 

%It is of interest to explore several promising directions to advance our research. 
%The focus lies on investigating alternative generative models, such as diffusion models, to overcome the known instability of GANs.
%Another interesting avenue is the expansion of the work of Coletta et al.~\cite{colettaRealisticMarketSimulations2021a} by incorporating order book level data, enabling applications in multi-stock simulation contexts.
%Lastly, we will modify the model's architecture and explore new features to better understand stock interdependence.
We believe this work demonstrates that a small GAN architecture can be effectively employed in specific domains, to preserve particular statistical properties of synthetic data. Most importantly, GANs' critics can be reused within more sophisticated diffusion models without the need for retraining. We release our code to foster future research in this direction.

\section*{Code and Data availability}
Data are extracted from \textit{LOBSTER}\footnote{https://lobsterdata.com/}, an online limit order-book data provider, which is publicly available for the research community with an annual fee.

We commit to publicly releasing our source code upon acceptance.

%In particular, future studies will analyze cascading behaviors across different stocks.
%This deeper understanding will improve simulation realism and offer valuable insights into market behavior dynamics.

%\backmatter

%\clearpage

\bibliographystyle{ACM-Reference-Format}
\bibliography{bio.bib}

%\appendix

\end{document}